\documentclass[10pt]{article} %

\newif\ifarxiv
\arxivtrue

\ifarxiv
  \usepackage[preprint]{tmlr}   %
  \renewcommand{\headrulewidth}{0pt}  %
\else
  \usepackage{tmlr}             %
\fi

\newcommand{\thesupp}{\ifarxiv the released materials (Appendix~\ref{app:artifacts})\else the supplementary material\fi}
\newcommand{\thesuppplain}{\ifarxiv the released materials\else the supplementary material\fi}

\usepackage{amsmath,amsfonts,bm}

\def\eqref#1{equation~\ref{#1}}

\def\1{\bm{1}}

\DeclareMathAlphabet{\mathsfit}{\encodingdefault}{\sfdefault}{m}{sl}
\SetMathAlphabet{\mathsfit}{bold}{\encodingdefault}{\sfdefault}{bx}{n}

\usepackage{hyperref}
\usepackage{url}

\ifarxiv
\hypersetup{pdftitle={What Roles Do System Prompts Really Assign? Validating and Steering Occupational Role Vectors in LLMs},
            pdfauthor={Jose L. Sampedro Mazon, Ilmi Yoon, Nadim Saad}}
\fi

\usepackage[utf8]{inputenc}
\usepackage{microtype}
\usepackage[dvipsnames]{xcolor}
\usepackage{booktabs}   %
\usepackage{amsmath}
\usepackage{amsfonts}
\usepackage{amssymb}
\usepackage{amsthm}
\usepackage{nicefrac}
\usepackage{algorithm}
\usepackage{algpseudocode}
\usepackage{subcaption}
\usepackage{graphicx}   %
\usepackage{wrapfig}
\usepackage{makecell}

\usepackage{pifont}

\newlength{\figcolwidth}
\providecommand{\toprule}{\hline}
\providecommand{\midrule}{\hline}
\providecommand{\bottomrule}{\hline}
\providecommand{\addlinespace}{\noalign{\vskip 2pt}}
\makeatletter
\let\resultsplaceholder\@undefined
\let\resultstable\@undefined
\let\resultstablecol\@undefined
\let\resultsfig\@undefined
\makeatother
\newcommand{\resultsplaceholder}[2]{%
  \fbox{\parbox{\dimexpr#1-2\fboxsep-2\fboxrule\relax}{\centering\small
  Missing file (upload the \texttt{figures} and \texttt{tables} folders):\\
  \footnotesize\detokenize{#2}}}}
\newcommand{\resultstable}[1]{%
  \IfFileExists{tables/#1.tex}%
    {\input{tables/#1}}%
    {\resultsplaceholder{\textwidth}{tables/#1.tex}}}
\newcommand{\resultstablecol}[1]{%
  \IfFileExists{tables/#1.tex}%
    {\input{tables/#1}}%
    {\resultsplaceholder{\figcolwidth}{tables/#1.tex}}}
\newcommand{\resultsfig}[2]{%
  \IfFileExists{#2}%
    {\includegraphics[width=#1]{#2}}%
    {\resultsplaceholder{#1}{#2}}}

\title{What Roles Do System Prompts Really Assign? Validating and Steering Occupational Role Vectors in LLMs}

\author{\centering
  \begin{tabular*}{\textwidth}{@{\extracolsep{\fill}}ccc@{}}
    Jose L.\ Sampedro Mazon & Ilmi Yoon & Nadim Saad \\
    {\normalfont\small jlsampedromazon@gmail.com} &
    {\normalfont\small i.yoon@northeastern.edu} &
    {\normalfont\small n.saad@northeastern.edu} \\
  \end{tabular*} \\[6pt]
  {\normalfont\itshape Khoury College of Computer Sciences, Northeastern University}}

\def\month{MM}  %
\def\year{YYYY} %
\def\openreview{\url{https://openreview.net/forum?id=XXXX}} %

\begin{document}

\maketitle

\begin{abstract}
Applications built on large language models assign the model a role through its system prompt, intending to shape how it responds for their use case. We introduce a validation framework for extracting and injecting occupational
role vectors, activation-level directions that capture how someone in a
given occupational role would distinctively respond to a task or question.
We validate these vectors through two complementary lenses, on
Qwen2.5-7B-Instruct and Llama-3.1-8B-Instruct. The interpretability lens
asks whether each vector is a stable, separable, domain-consistent
representation of its role: extracted vectors are highly reproducible
across repeated runs and sampling temperatures, remain separable even for
very similar roles, and agree more with themselves across in-domain and out-of-domain questions than with other roles' signal. The control
lens asks whether injecting a role vector during generation makes the
model express the role, and sustain it over a conversation, as well as a
role prompt does. On Qwen, injecting the vector alone makes the model express the intended role substantially more often than unsteered, but at best matches a role prompt; on both models, adding the vector to the prompt strengthens the role in single answers but did not slow the role's decline over 30 turns. Together, the two lenses show why both are needed: out of domain the role is expressed only subtly, yet the activations still encode it; and a vector that is stable and separable can still barely steer the model's behavior.
\end{abstract}

\section{Introduction}
\label{sec:introduction}

Applications built on large language models (LLMs) assign the model a role through a system prompt, instructing it how to present itself and respond before any user turn \citep{zheng2024}. In an analysis of 1,309 system prompts, from general-purpose products such as ChatGPT and Claude to user-built chatbots for single use cases, role and identity instructions are present in over 90\% \citep{neumann2026controls}. However precisely the system prompt specifies the role, how it may change the model’s response cannot be inferred from the prompt alone \citep{neumann2026controls}. The system prompt is ultimately an input; its effect is determined by the model’s computation over that input, and becomes observable only in the model's outputs or in its activations. The activations that carry the role are rarely examined. Recent work does describe personas at activation level, representing character traits and the default assistant identity as directions that can be read to detect a trait before it shows in outputs and written to suppress it \citep{zou2023,chen2025,lu2026}. In that work a direction counts as valid when it predicts or steers the behavior it names: the test is the effect. We ask the preceding question, whether the direction is a reliable representation at all, stable across extractions, separable from its neighbors and consistent across domains; and then whether a direction that passes those tests also controls behavior, which the effect-based test leaves open. We ask both for occupational roles assigned through a system prompt, which that work does not represent.\footnote{We used AI assistants (Claude, Anthropic) to help write and edit the text, format the \LaTeX{} source, check citations against their sources, and write and debug code. The authors designed the study, ran and checked the experiments, and take full responsibility for all ideas, claims and results.}

An application that relies on an assigned role needs three things of the model: stability, separability from closely related roles, and persistence. Stability means the role the model takes is the same across independent runs and sampling temperatures, not an artifact of one stochastic generation. Separability means the model holds the role’s nuanced stance, not only the subject matter it shares with its neighbors. Persistence means the role holds throughout an interaction, across every turn of the application’s flow. Separability is hardest to assess when roles share knowledge: a mergers-and-acquisitions (M\&A) lawyer and a corporations law professor can answer many questions alike, so their answers may not reveal whether the model represents them differently. Persistence is difficult to achieve through prompting alone, since system-prompt instructions can drift within eight rounds of conversation \citep{li2024}. We choose occupational roles because they are common in deployed prompts and in multi-agent frameworks that give each agent an occupation \citep{hong2024}, and because closely related occupations share knowledge, which makes separability testable.

Role instructions in the system prompt are one of three levels at which a role can be assigned: the model's input, its weights, or its internal activations. Prompting acts on the input. Fine-tuning acts on the weights and requires training data collected for each role \citep{shao2023}. Representation engineering acts on the activations, treating a role as a direction in activation space \citep{zou2023,chen2025,lu2026}. We call this direction a role vector and obtain it by contrast: the activations the model computes under the role's system prompt, minus those it computes on the same input under its default prompt, with no role assigned. Because it is the difference the system prompt makes, the vector can be examined as a representation of the assigned role, or injected during generation to assign the role without the prompt. The first is a question of interpretability, the second of control. Injection can steer traits without training \citep{turner2023,chen2025}. Whether a direction's geometry predicts whether it steers has been asked across many behaviors, with conflicting answers \citep{braun2025,bas2025}; it has not been asked of a direction validated as the representation of one role, tested against the prompt that assigns it.

We study these questions for the two training-free ways to assign a role, prompting and activation steering, through two lenses. The interpretability lens examines the role vectors themselves, asking whether each is a valid representation of its role in the sense of construct validation \citep{cronbach1955,jacobs2021}: stable across repeated extractions and sampling temperatures, separable from closely related roles, and consistent whether built from questions inside or outside the role's domain. The control lens examines the model's behavior, asking whether injecting a role vector makes the model express the role in single answers, and sustain it over an extended interaction, as well as a role prompt does. An LLM judge scores every answer; an answer succeeds when it is scored fully in role and coherent, and the \emph{success rate} is the share of answers that succeed. Together, the two lenses separate what the model represents from what it expresses.
\\
\noindent
In this work, we make the following contributions:

\begin{itemize}
  \item We introduce an extraction protocol that separates occupational
  role identity from question content by having every role answer the same
  questions from all roles' domains, contrasted against the model's default
  activations on the same question. The resulting role vectors on
  Qwen2.5-7B-Instruct and Llama-3.1-8B-Instruct are reproducible, separable
  even for closely related roles, and consistent across question domains.

  \item We find that injecting a role vector makes the model express that specific role, strongly on Qwen and weakly on Llama, but on either model no more clearly than role prompting alone. Vector steering alone raises the success rate over the unsteered model by up to $+22.8$ percentage points on Qwen and up to $+7.0$ on Llama. Against the role prompt, however, steering alone at best matches it on Qwen and trails it on Llama by about 20 percentage points. We also find that vector steering at low strength combined with role prompting yields the strongest role expression of the conditions tested, raising the success rate over the prompt alone by $+5.0$ to $+7.8$ percentage points wherever the combination stayed coherent.

  \item We find that steering with a role vector does not make a role more persistent than prompting. Over 30-turn conversations on Qwen, a role injected alone declines at a rate not reliably different from a prompted role, and the combination of vector and prompt gives a stronger start that fades to the prompted level. On Llama, a role injected alone stays below the prompted role throughout, and the combination declines at the prompted rate, with a lead only in the last five turns that we treat as suggestive.

  \item We show that the interpretability and control lenses answer
  different questions, and that both are needed. Out of domain, the control lens finds the model's responses rarely express a role fully, as expected where a role should show only subtly, yet the interpretability lens shows the role still present in activations on those questions. Conversely, Llama's vectors pass the same activation-level tests as Qwen's, yet barely change Llama's behavior under the control lens, so passing activation-level tests does not by itself establish that a vector can be used for control.
\end{itemize}

\noindent Section~\ref{sec:related-work} reviews how prompted roles shape and lose their hold on behavior, how a role can be specified through prompts, fine-tuned into the weights or steered at the activation level, and how assigned roles are evaluated.

\section{Related Work}
\label{sec:related-work}

\paragraph{Roles assigned through prompts.}
A role assigned through a prompt measurably shapes behavior, altering toxicity and bias \citep{wan2023,wangS2025}. A
role held through prompting alone tends to degrade. Instructions drift over
dialogue turns \citep{li2024}. Post-training only loosely holds models near
a default assistant region of activation space, and models drift from it
\citep{lu2026}. A model's own built-in leanings can also pull it away from
an assigned stance \citep{taubenfeld2024}. Most of these studies, like deployed applications, judge whether a role is held by what the model outputs.

\paragraph{Ways to assign a role.}
Roles can be specified through natural-language prompts \citep{tseng2024},
or embedded parametrically through fine-tuning \citep{shao2023,abdulhai2025},
which demands data collected for each role.
Representation engineering instead reads and writes behavioral properties
as directions in activation space \citep{zou2023}. Post-trained models
carry a default assistant identity occupying its own region of this space
\citep{lu2026}, and contrasting a role-prompted model's activations with
the default yields an interpretable direction for that role, though role
directions overlap substantially with one another \citep{lu2026}. Adding
such a direction to the residual stream modifies behavior without
retraining \citep{turner2023,rimsky2024,chen2025}. \citet{poterti2025} build role vectors by contrasting role-specific prompts
with generic ones rather than with a default assistant on the same
questions, and check role identity only by asking the model what a token
carrying the vector means; neither role expression in generated answers
nor the role prompt is tested. \citet{wang2025roleplaysteering} also start from the difference a role prompt
makes to activations, but build the steering vector from sparse-autoencoder
features (a decomposition of activations into many more interpretable
directions) rather than from the raw activation difference. They use two
roles, such as a mathematics teacher, to improve reasoning, and score steering
by the share of arithmetic and commonsense questions answered correctly rather
than by how closely answers adhere to the role; on that measure it outperforms
the role prompt in most settings tested. The vector's stability and separation
from other roles, and the role's persistence over a conversation, are not
tested.

Prompting beats steering on concept-insertion targets \citep{wu2025axbench}
and, in single answers, on persona archetypes including occupations
\citep{bas2025}; steering effects vary across inputs and behaviors
\citep{tan2024} and models \citep{dasilva2025}, and residual-stream steering
degrades over ten turns \citep{kang2026}. Whether a direction's geometry
predicts control is disputed: \citet{braun2025} find that coherent,
well-separated directions steer better, whereas \citet{bas2025} find that
separation does not predict steering success.

\paragraph{Evaluation and validation of roles.}
Role fidelity, steering, and drift are evaluated from outputs: whether
answers match the intended character \citep{wangX2024,samuel2025}, how far
behavior shifts along a target trait \citep{turner2023,chen2025}, and how
much behavior changes over dialogue turns \citep{li2024}. These judgments
can come from LLM judges, which match human preferences on broad
comparisons \citep{zheng2023}, vary in reliability by task
\citep{bavaresco2025}, and fall short of humans at identifying which
character is speaking \citep{zhou2025personaeval}. Output-based evaluation
shows a model behaves consistently with a role, not that an extracted
vector reliably represents it. Behavioral consistency can coexist with
vectors unstable across extractions, entangled with nearby roles, or
dominated by domain content, faults that surface only in use. Measurement
theory treats these as separate questions, reliability and construct
validity \citep{cronbach1955,campbell1959convergent,jacobs2021,wallach2025}.
Probing work adds a converse caution: a probed property need not be stored
in activations \citep{hewitt2019} or used by the model \citep{ravichander2021},
and an intervention can work through a pathway the model does not normally
use \citep{makelov2024}.

\section{Extracting and Injecting Occupational Role Vectors}
\label{sec:extraction-injection}

Section~\ref{sec:extraction} extracts role vectors and tests at the
activation level whether they are stable, separable and domain-consistent;
Section~\ref{sec:injection} injects them and tests at the behavioral level
whether the role is expressed in single answers and persists over a
conversation.

\subsection{Extraction of Occupational Role Vectors}
\label{sec:extraction}

\subsubsection{Eliciting Role-Specific Activations}
\label{sec:eliciting}

We generate role-conditioned responses by directly prompting the model to adopt a
given role, specifying the occupational habits and characteristic reasoning that
should manifest in both domains. Every role answers the same fixed set of
questions, ensuring that any differences we observe reflect the role itself rather than the questions posed to each role. This set includes questions from
the role's own domain, which we call in-domain, and questions from other roles'
domains, which we call out-of-domain. If a role's distinctive behavior only shows
up on questions from its own domain, we cannot tell whether that behavior comes
from the role or from the subject matter of those questions
\citep{cronbach1955,campbell1959convergent}. The full question set appears in \thesupp.

We score only the first-turn response, as prior work shows that behavior set by a system prompt can degrade over the course of multi-turn dialogue,
drifting toward the model's default behavior or being overridden by inherited
priors \citep{li2024,taubenfeld2024}. We grade each response using an LLM judge
on a role-specific, domain-aware rubric, following the same approach used to
grade character trait expression in generated text \citep{chen2025}. The rubric
scores each response from 0 to 3, with higher scores indicating stronger, more
natural role expression. We separately score each response's coherence on a 0 to
100 scale, using a fixed rubric applied uniformly across roles. The judge is gpt-4.1-mini (median of three trials), and a blind human check of its scores is reported in Appendix~\ref{app:ood-scoring} (Table~\ref{tab:M1}); the full rubric appears in \thesupp.

A response contributes to the activation sample only if it meets three
conditions: its adherence score is at least $s_{min}$, its coherence score is at
least a fixed threshold, and it is not degenerate, meaning it does not contain
excessive repeated phrases or non-ASCII characters. For each role $r$, question
$i$, layer $l$, and run $s$, we extract the model's activation
$h_{r,i,l,s} \in \mathbb{R}^d$, where $d$ is the model's hidden dimension,
together with its adherence score, across all layers. In
Section~\ref{sec:validating-extraction}, we provide a framework to test that
this activation extraction process yields a reproducible representation of the
role across repeated runs.

\subsubsection{Building Role Vectors from Activations}
\label{sec:building-vectors}

We construct each role's vector $v_{r,l,s}$ starting from the scored activations
$h_{r,i,l,s}$ for role $r$, question $i$, layer $l$, in three steps.

\begin{enumerate}
  \item \textbf{Subtract the default-prompt activation.} For each question
  $i$, we subtract the activation recorded when the model answers that same
  question under its default prompt from the role's activation,
  \begin{equation}
  \tilde{h}_{r,i,l,s} = h_{r,i,l,s} - h_{default,i,l,s}
  \label{eq:subtraction}
  \end{equation}
  Equation~(\ref{eq:subtraction}) isolates the contribution of the role prompt
  itself from the content of the question and the model's default response
  tendencies, both of which are otherwise present in $h_{r,i,l,s}$ regardless
  of role.

  \item \textbf{Weight by score.} Each response with a score at or above
  $s_{min} = 2$ contributes to the vector in proportion to its score. Score-2
  responses are common out-of-domain, where role expression often appears
  through subtler habits rather than explicit occupational reasoning.

  \item \textbf{Balance in-domain and out-of-domain.} We compute an
  in-domain-only vector $v_{in,r,l,s}$ and an out-of-domain-only vector
  $v_{out,r,l,s}$, each following Steps 1 and 2, and average them with equal
  weight:
  \begin{equation}
  v_{cand,r,l,s} = 0.5 \cdot v_{in,r,l,s} + 0.5 \cdot v_{out,r,l,s}
  \label{eq:balancing}
  \end{equation}
  Out-of-domain questions make up 83\% of the full set, versus 17\% in-domain.
  Averaging the two domains equally, rather than pooling all questions
  together, prevents the larger out-of-domain share from dominating the final
  vector.
\end{enumerate}

\subsubsection{Validating Extraction: Stability, Separability, and Domain
Consistency}
\label{sec:validating-extraction}

\textbf{Stability.} We compute the cosine between every pair of a role's
repeated extractions, and between the centroids of two random halves of
them, repeated over many splits so the result does not hinge on a few
unusual extractions. To check that this agreement is specific to the role,
we test whether each extraction is closer to its own role's centroid than
to any other, and compute each role's margin over the nearest other role
with a bootstrap interval. We also compare each role's centroid across
sampling temperatures.

\textbf{Separability.} We measure the angle between every pair of role
centroids, focusing on two pairs designed to be hard to separate: a
corporate M\&A lawyer and a corporations law professor share a legal
domain but differ in professional stance, while a general professor and
the corporations law professor share an academic stance but differ in
specialization.

\textbf{Domain consistency.} For each extraction we build one vector from
in-domain questions only and one from out-of-domain questions only. If the
role signal is genuine, same-role vectors should agree more than
different-role vectors, and each half should identify its role using only
the other half's centroids. We also check that the balanced vector of
Equation~(\ref{eq:balancing}) does not lean toward either half.

\subsection{Injection of Occupational Role Vectors}
\label{sec:injection}

\subsubsection{Steering Answer Generation}
\label{sec:steering}

During generation, we add the role vector to the model's residual stream at
the injection layer, at every step after the prompt. The prompt itself is
processed unsteered, so the model's first generated token is already a
response to the unsteered prompt, and only the tokens generated after it are
steered.

At the injection layer, we rescale the vector to a fixed size relative to the
model's own typical activation size at that layer, rather than injecting it
at a fixed raw magnitude, so the injected strength is comparable across
layers and models. When the injection site spans more than one layer, each
additional layer adds only the difference needed to reach its own target
size. The injection site and strength are chosen by calibration on a
held-out question set, under a rule fixed before seeing the results, with the full procedure in \thesupp.

We compare five conditions: unsteered, prompted (role given through a
system prompt only), steered (the role vector added to the activations, no
prompt), prompt plus vector (both together), and a random-direction
control, which injects a random vector of the same size in place of the
role vector to test whether any effect comes from the specific role
direction rather than from adding any vector of similar size.

\subsubsection{Validating Injection: Expression and Persistence}
\label{sec:validating-injection}

\textbf{Expression.} We test whether a single answer reflects the injected
role, comparing success rate across the conditions from
Section~\ref{sec:steering}. An answer succeeds if it is judged as adherent
to the role, coherent, and free of degenerate text, using the same judging
criteria as extraction. Success rising under injection could still reflect
the model writing more fluently or more confidently overall, rather than
expressing the specific injected role, so we rule this out in two ways. We
score each answer against all six roles' rubrics and confirm that the rise
in success under injection is concentrated on the role's own rubric rather
than spread evenly across all six. We also ask a judge, not told which role
was intended, to identify which of the six roles the answer sounds like,
testing recognizability independently of the rubric-based scoring in the
first check.
\\
\noindent
\textbf{Persistence.} We test whether the injected role holds up over an
extended conversation, as well as a prompted role does. The primary test
looks at the sign of the slope of judged role adherence across turns, which
we call the decline rate, compared between the injected condition and the
prompted baseline. A negative decline rate means the role decays faster
under injection than under prompting. A secondary check compares mean
success over a window of turns, injected condition minus prompted, rather
than the decline rate across the whole conversation. This can catch a
persistent gap even when the decline rate does not differ. Each model is run at one temperature, chosen by a rule fixed in advance as
the one with the largest single-turn steering gain whose interval excludes
zero: Qwen $T=1.0$, Llama $T=0.0$. The same rule selects the settings for
role identification.

\section{Results}
\label{sec:results}
 
We report validation results at activation-level first, then behavioral. Both are measured at the injection site (Qwen layers 19--21, Llama layer 12), so one vector is validated at both levels. Differences are in percentage
points (pp). Behavioral results compare \emph{conditions}, the ways we run the model when
it answers (e.g. unsteered, steered, prompted), always within
one \emph{setting}, a model at one temperature. Every comparison has a 95\%
interval (question-cluster bootstrap for single answers; conversation
bootstrap over 18 conversations of 30 turns, with a Wilcoxon signed-rank test)
given in the tables and figures; in the text, we report an effect only when
its 95\% interval excludes zero, and otherwise say the two conditions are not
distinguishable.

\subsection{Activation-level validation}
\label{sec:results-geometry}
 
\noindent\textbf{The role direction is a stable signal: 30 independent
extractions agree, at every temperature.}
We extract each role 30 times in independent runs and average the results
into a centroid. If sampling noise dominated, the 30 extractions would scatter
and the centroid would change direction with temperature. Neither occurs
(Table~\ref{tab:A1}). The 30 extractions of a role agree closely at every
temperature above 0, where sampling varies the answers: their same-role
pairwise cosine is 0.898 to 0.984 on Qwen and 0.907 to 0.969 on Llama (at
$T=0.0$ the extractions are near-copies, 0.995 or above), and averages built
from two random halves of the 30 extractions agree at cosine 0.993 or above on
both models. Each role's centroid also stays aligned with its $T=0.0$
centroid at cosine 0.954 or above on Qwen and 0.950 on Llama. These results are reported at each model's injection site; the same
stability measures at the pilot reference layers and the most separable
layer band are similar, and the full separability and cross-domain tables
appear in Appendix~\ref{app:layers}.
 
\begin{table}[t]
\centering
{\footnotesize\setlength{\tabcolsep}{3pt}
\begin{tabular*}{\columnwidth}{@{}l@{\extracolsep{\fill}}ccccc@{}}
\toprule
Model & & $T$ &
\begin{tabular}[b]{@{}c@{}}Pairwise\\cosine\end{tabular} &
\begin{tabular}[b]{@{}c@{}}Split-half\\cosine\end{tabular} &
\begin{tabular}[b]{@{}c@{}}Cosine to\\$T{=}0$ centroid\end{tabular} \\
\midrule
Qwen  & & 0.0 & 0.997--1.000 & 0.9998 & 1     \\
      & & 0.5 & 0.938--0.984 & 0.9957 & 0.976 \\
      & & 1.0 & 0.915--0.977 & 0.9939 & 0.971 \\
      & & 1.5 & 0.898--0.964 & 0.9927 & 0.954 \\
\midrule
Llama & & 0.0 & 0.995--0.998 & 0.9997 & 1     \\
      & & 0.5 & 0.930--0.969 & 0.9950 & 0.971 \\
      & & 1.0 & 0.907--0.954 & 0.9932 & 0.950 \\
\bottomrule
\end{tabular*}}

\caption{Stability of the role vectors at the injection site (Qwen layers
19 to 21; Llama layer 12), 30 extractions per role, six roles. Pairwise
cosine: mean cosine between every two of a role's extractions, lowest and
highest over roles (and Qwen layers). Split-half cosine: mean over 200
random half splits of the cosine between the two half centroids, lowest
over roles and layers. Cosine to $T=0.0$ centroid: lowest over roles and
layers (1 at $T=0.0$ by definition). Leave-one-run-out assignment is
100\% in every setting.}
\label{tab:A1}
\end{table}
 
\noindent\textbf{Roles are separable, including the two closest occupational
pairs.}
The roles include two designed close pairs: an M\&A lawyer and a corporations
law professor (same field, corporate law; different stance), and a general
professor and the law professor (same stance; generalist vs specialist field).
At $T=1.0$ the \emph{angular separation} (the angle between two roles'
centroids) is 48.8--53.1$^{\circ}$ on Qwen (layers 19 to 21) and
37.1$^{\circ}$ on Llama for the first pair, and 54.4--58.1$^{\circ}$ and
41.9$^{\circ}$ for the second (appendix Table~\ref{tab:B1}). The \emph{nearest-role margin} (an extraction's cosine to its own role's
centroid minus that to the nearest other role's centroid, averaged over the
role's 30 extractions) is clearly positive at $T=1.0$ for the three roles in
the designed pairs: 0.337 to 0.397 on Qwen and 0.197 to 0.244 on Llama
(appendix Table~\ref{tab:B1}).
Every extraction is also closest to its own role's centroid when left out of it (100\%; Table~\ref{tab:A1}). Llama's centroids are less widely separated than Qwen's, but its
extractions agree as closely and every one is still assigned to its own
role.
 
\noindent\textbf{The role vector is consistent across in-domain and
out-of-domain answers.}
Each extraction's vector is built from all 90 questions, 15 in-domain and 75
out-of-domain, with the two sets weighted equally. Two checks ask whether it points the same way whichever set it is built from. The first compares, within each extraction and at each
temperature, an \emph{in-domain vector} (built from the 15 questions) with an
\emph{out-of-domain vector} (built from the 75; appendix Table~\ref{tab:D1},
Figure~\ref{fig:D2}). For the same role these two vectors are clearly more
aligned in direction than for different roles: at $T=1.0$, mean cosine 0.79 vs
0.37 on Qwen and 0.77 vs 0.55 on Llama (180 same-role and 900 different-role
vector pairs, each pair from one extraction; Qwen at its least favorable
layer). Matching each in-domain vector to the nearest average of the out-of-domain
vectors, and the reverse, gives the right role at every temperature, except
for the Llama cashier's out-of-domain vectors, matched correctly in as few as
10\% of its 30 extractions ($T=0.5$; appendix Table~\ref{tab:D1}). The second check asks whether the role's centroid, the vector we later inject,
leans towards either question set. We compare it with the average of the
role's 30 in-domain vectors and with the average of its 30 out-of-domain
vectors. At $T=1.0$ it points in a similar direction to both: cosine 0.96 to
the in-domain average on both models (lowest role 0.94 on Qwen, 0.92 on Llama)
and 0.96 on Qwen and 0.97 on Llama to the out-of-domain average (lowest role
0.92 and 0.94). Either average still assigns every extraction to its own role and separates
the close pairs only slightly more (Table~\ref{tab:C1}).
 
\subsection{Behavioral validation}
\label{sec:results-behavior}
 
\noindent\textbf{Injecting the role vector makes the model express that
specific role: strongly on Qwen, weakly on Llama.}
We inject the vector at a \emph{strength} chosen on calibration questions. On Qwen at $T=1.0$ and strength 0.3, steering raises the success rate from
9.3\% to 32.0\%, a \emph{steering gain} of $+22.8$~pp (540 answer pairs per
setting, pooled over the six roles), with similar gains at $T=0.0$ and $0.5$
(Table~\ref{tab:E1}).\footnote{Absolute success rates are low by design: five sixths of the
answers (450 of 540) are to out-of-domain questions, where full adherence
requires expressions a neutral assistant would not produce, so we compare
conditions rather than read levels. Appendix~\ref{app:ood-scoring} shows
the judge still separates conditions on out-of-domain answers.} At
$T=1.5$ maintaining coherence limits the steering strength to 0.1 and the gain
is not distinguishable from zero. 

Three checks show the gain belongs to the steered role rather than to a
general change in style. \emph{Role specificity}, the rise in success on
the steered role's own rubric minus the mean rise on the other five, is
$+24.7$~pp on Qwen at $T=1.0$ (Figure~\ref{fig:E2}); a judge choosing among
all six roles identifies the steered role in 36.9\% of answers against
17.0\% unsteered (chance 16.7\%; Table~\ref{tab:BID}); and a random
direction of the same size leaves success not distinguishable from
unsteered at any temperature (Table~\ref{tab:E1}). On Llama, steering
works only at $T=0.0$ ($+7.0$~pp; specificity $+9.0$~pp; identification
26.5\% vs 17.2\%, out of domain only to chance) and is not
distinguishable from zero at $T=0.5$ and $1.0$, although its vectors are
stable and separable.

\begin{table*}[t]
\centering
{\footnotesize\setlength{\tabcolsep}{3pt}
\begin{tabular}{llrrrrr}
\toprule
Model & T & \shortstack[l]{Steered\\ strength} & \shortstack[l]{Steered minus\\ unsteered} & \shortstack[l]{Random direction\\ minus unsteered} & \shortstack[l]{Specificity,\\ all six rubrics} & \shortstack[l]{Specificity, without\\ general professor} \\
\midrule
Qwen & 0.0 & 0.3 & +16.3 [+12.2, +20.6] & +0.9 [$-$1.9, +3.7] & +18.2 [+14.7, +21.7] & +19.7 [+16.0, +23.5] \\
Qwen & 0.5 & 0.3 & +19.1 [+14.8, +23.3] & +0.7 [$-$1.7, +3.1] & +19.4 [+16.2, +22.7] & +21.7 [+18.3, +25.1] \\
Qwen & 1.0 & 0.3 & +22.8 [+18.5, +27.0] & +2.4 [$-$0.6, +5.4] & +24.7 [+20.8, +28.4] & +25.1 [+21.0, +29.1] \\
Qwen & 1.5 & 0.1 & +3.1 [$-$0.7, +7.2] & +0.2 [$-$3.0, +3.3] & +3.5 [+0.6, +6.4] & +3.2 [+0.4, +6.0] \\
Llama & 0.0 & 0.5 & +7.0 [+4.4, +9.4] & $-$0.9 [$-$3.5, +1.5] & +9.0 [+6.9, +11.1] & +5.2 [+3.1, +7.4] \\
Llama & 0.5 & 0.2 & +1.1 [$-$1.7, +3.7] & $-$1.9 [$-$4.6, +0.7] & +2.4 [+0.2, +4.7] & +1.0 [$-$1.0, +3.2] \\
Llama & 1.0 & 0.2 & +0.2 [$-$3.0, +3.5] & $-$1.1 [$-$3.5, +1.3] & +4.1 [+1.9, +6.4] & +2.0 [$-$0.0, +4.1] \\
\bottomrule
\end{tabular}}

\caption{Steering gain and controls for every model and temperature, 540
answer pairs per setting: steered minus unsteered and random direction minus
unsteered success rate (pp, 95\% question-cluster intervals in brackets), and
role specificity (gain on the steered role's own rubric minus the mean gain on
the other five rubrics, pp, 95\% intervals), with all six rubrics and without
the general professor rubric. Steered strength in reference-norm units.}
\label{tab:E1}
\end{table*}
 
\begin{figure}[t]
\centering
\resultsfig{\linewidth}{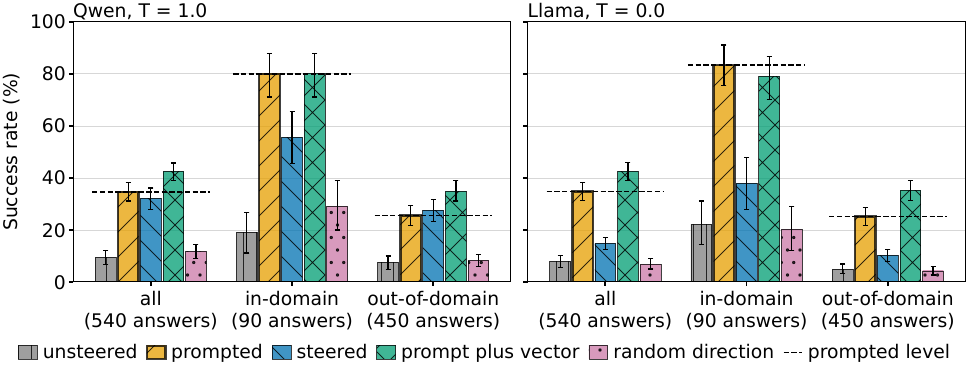}
\caption{Success rate by condition over all 540 answers, the 90 in-domain
answers and the 450 out-of-domain answers, 95\% question-cluster intervals;
dashed line: the prompted level. Left: Qwen, $T=1.0$ (steered 0.3, prompt
plus vector 0.15). Right: Llama, $T=0.0$ (steered 0.5, prompt plus vector
0.25). An answer succeeds if judged adherence 3 of 3, coherence at least 75 of
100, and not degenerate.}
\label{fig:F1}
\end{figure}

\noindent\textbf{In single answers, steering alone does not express the role
more clearly than a role prompt.}
On Qwen at $T=1.0$, the steered minus prompted success rate is $-2.6$~pp
overall and $+1.8$~pp out of domain (27.3\% vs 25.6\%), neither
distinguishable from zero, but $-24.4$~pp in domain (55.6\% vs 80.0\%;
Figure~\ref{fig:F1}; appendix Table~\ref{tab:F2}). Steering thus never exceeds
prompting; the pooled difference is small because out-of-domain questions
make up five sixths of the answers. One possible reason for the in-domain gap,
which we do not test, is that the role prompt spells out how the role reasons
about its own domain. At $T=0.0$ and $0.5$ the prompt also leads overall
($-10.9$ and $-8.5$~pp), mostly in domain. Role identification likewise finds
no difference at $T=1.0$: 36.9\% of steered and 37.8\% of prompted answers are
identified overall and 25.3\% vs 26.4\% out of domain, with both at 94.4\% in
domain (Table~\ref{tab:BID}). On Llama the prompt leads in every setting
(about $-19$ to $-20$~pp) and in role identification ($T=0.0$: 26.5\% vs
55.9\%).
 
\noindent\textbf{A role-prompted model steered with the role vector at low strength expresses the role more clearly in single answers.}
Steering a role-prompted model at low strength (0.15; 0.25 for Llama at
$T=0.0$) raises the success rate over the prompt alone by $+5.0$ to
$+7.8$~pp in all five settings where a strength kept coherence within the
floor (appendix Table~\ref{tab:F2}). The gain comes from out-of-domain
questions, where role specificity also rises ($+5.9$ to $+11.6$~pp), so it
is not a general effect of longer answers; on Llama it is partly driven
by length and only weakly supported by role identification
(Appendix~\ref{app:ppv}). The cost is coherence: on Qwen at $T=1.0$, mean
coherence falls from 95.3 to 91.7.
 
\begin{figure}[t]
\centering
\resultsfig{\linewidth}{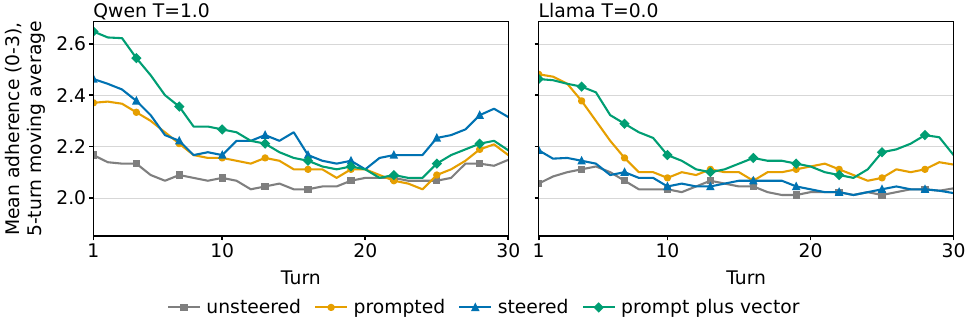}
\caption{Mean judged adherence (0--3) per turn, 5-turn moving average, 18
conversations per condition; the vertical axis spans 1.85 to 2.69, not 0
to 3. Left: Qwen, $T=1.0$. Right: Llama, $T=0.0$.}
\label{fig:H1}
\end{figure}

\noindent\textbf{Over 30 turns, steering does not make the role more
persistent than prompting.}
Persistence is measured by the \emph{decline rate}, the per-conversation slope
of adherence over turns, compared with the prompted condition on the same 18
conversations (\emph{slope difference}, Wilcoxon signed-rank $p$;
Figure~\ref{fig:H1}, Table~\ref{tab:H2}); a secondary \emph{level} comparison
uses mean success over a window of turns (appendix Table~\ref{tab:H2b}). On Qwen at $T=1.0$, every condition
declines. Steering alone declines at a rate not distinguishable from prompting
($p=0.071$); prompt plus vector starts highest but declines faster
($p=0.006$) and ends at the prompted level. Over turns 26 to 30, steering
alone leads prompting by $+13.3$~pp ($p=0.044$); since this window was chosen
after the first results, we treat the lead as suggestive. This late-turn advantage holds under several robustness checks (excluding
one role, excluding turns cut off at the token limit, and restricting to
conversations never used for calibration), detailed in
Appendix~\ref{app:mt-robust}. On Llama at $T=0.0$, steering
alone stays below prompting over turns 1 to 30 ($-8.5$~pp); its slower
decline ($p=0.049$) starts from 11.1\% vs 44.4\% success at turn 1, so it
likely reflects less room to fall rather than better persistence. Prompt plus
vector declines at the prompted rate ($p=0.605$) and leads over turns 26 to
30 ($+11.1$~pp, $p=0.047$), also suggestive. The kept-history condition, which
reuses the key-value cache computed while steering earlier answers, declines
at a rate not distinguishable from the steered condition on either model
($p=0.66$ Qwen, $p=0.33$ Llama).
 
\begin{table*}[t]
\centering
{\small\setlength{\tabcolsep}{4pt}
\begin{tabular}{lllrr}
\toprule
Model & \shortstack[l]{Condition\\ (strength)} & \shortstack[l]{Slope difference,\\ adherence per turn} & \shortstack[l]{Adherence,\\ turns 1-5} & \shortstack[l]{Adherence,\\ turns 26-30} \\
\midrule
\multicolumn{5}{l}{\textit{Qwen2.5-7B-Instruct, T=1.0 (18 conversations per condition)}} \\
Qwen & steered (completion, 0.25) & +0.0037 [$-$0.0013, +0.0087], p=0.071 & 2.42 & 2.32 \\
Qwen & prompt plus vector (0.15) & $-$0.0083 [$-$0.0132, $-$0.0035], p=0.006 & 2.62 & 2.21 \\
Qwen & kept history (0.25) & +0.0027 [$-$0.0030, +0.0079], p=0.304 & 2.39 & 2.27 \\
Qwen & prompted (reference) & reference & 2.37 & 2.19 \\
\midrule
\multicolumn{5}{l}{\textit{Llama-3.1-8B-Instruct, T=0.0 (18 conversations per condition)}} \\
Llama & steered (completion, 0.15) & +0.0050 [+0.0008, +0.0092], p=0.049 & 2.16 & 2.03 \\
Llama & prompt plus vector (0.15) & +0.0011 [$-$0.0026, +0.0047], p=0.605 & 2.44 & 2.24 \\
Llama & kept history (0.15) & $-$0.0005 [$-$0.0108, +0.0083], p=0.420 & 2.18 & 1.91 \\
Llama & prompted (reference) & reference & 2.44 & 2.11 \\
\bottomrule
\end{tabular}}

\caption{Decline rate against the prompted condition (adherence slope per
turn, 95\% conversation bootstrap interval, tie-robust Wilcoxon $p$) and
mean adherence over the first and last five turns. Level differences:
appendix Table~\ref{tab:H2b}.}
\label{tab:H2}
\end{table*}
 
\noindent\textbf{Stable, separable vectors do not guarantee steering.}
In each setting, the steering strength is the most successful calibration
strength whose mean coherence stays above the coherence floor
(Figures~\ref{fig:G2}, \ref{fig:I1}). On Llama this gives a steering gain only
under greedy decoding (strength 0.5 at $T=0.0$); at $T=0.5$ and $1.0$ the
selected strength is 0.2 and the steering gain is not distinguishable from
zero. Even at 0.2, coherence suffers: at $T=1.0$, 8.3\% of steered answers run
into the 512-token length limit (unsteered 1.1\%) and mean coherence falls
from 96.0 to 85.7. In an exploratory sweep over further injection sites and strengths,
Llama's mean adherence never exceeds the unsteered level (0 of 103
combinations, against 37 of 70 on Qwen; Appendix~\ref{app:sweep}). As a control, a language vector built and injected the same way steers
Qwen into clean, coherent French but fails on Llama at every strength, so
the injection code works and Llama's failure is not specific to roles
(Appendix~\ref{app:french}). Nor
is it the geometry: Llama's vectors are stable and clearly separable
(Section~\ref{sec:results-geometry}).
 
A large share of each role vector (36 to 84\% of its squared length) lies
along a direction shared by all six roles, yet on Qwen at $T=1.0$ the
role-specific remainder alone gives $+16.9$~pp of the whole vector's
$+22.8$~pp steering gain, against $+4.4$~pp for the shared part
(Appendix~\ref{app:shared}).
 
\noindent\textbf{Activation-level and behavioral validation answer different
questions, and the protocol needs both.}
Behavior alone cannot locate a failure: Llama's vectors are stable and separable yet barely steer, which rules out the geometry we measured as the cause; without the geometry, a malformed vector and a model that resists steering would look the same. Nor can behavior alone tell a role that is
absent from one that is present but not clearly expressed: out of domain,
success is at most 35\% in every condition, yet out-of-domain vectors still
point to their own role for every role on Qwen and five of six on Llama
(appendix Table~\ref{tab:D1}). Activation-level validation shows whether the role is
present in the model; behavioral validation shows whether it reaches the
output.

\section{Conclusions}

We validated occupational role vectors in two 7--8B models through two
lenses. Through the interpretability lens, the vectors are stable,
separable even for closely related roles, and consistent across question
domains. Through the control lens, injecting a vector makes Qwen express
its role ($+22.8$~pp at $T=1.0$) but no more clearly than a role prompt,
and it does not slow the role's decline over 30 turns; vector steering at low strength in combination with role prompting strengthens the role in single answers. Llama's vectors are stable and separable yet barely steer. Since roles assigned through system prompts are rarely inspected inside the model, a role should be validated at both levels, for product reliability and for AI safety alike, before steering is relied on for control.

\section*{Limitations}

\paragraph{Scope.}
We study two instruction-tuned models of 7 to 8 billion parameters,
six benign occupational roles in English, and one injection site per
model (Qwen layers 19 to 21, Llama layer 12), with the vector added
only on generated tokens; no harmful behavior was tested. Other
scales, non-occupational roles, other languages, and the tool-use and
multi-agent settings where roles are assigned in practice remain to be
tested.

\paragraph{Judging.}
All scores come from one judge (gpt-4.1-mini-2025-04-14; median of
three trials for adherence and coherence), and role identification uses the same model and shows
it each role's rubric, so the measures are not independent; role
identification was run at one temperature per model (Qwen $T=1.0$,
Llama $T=0.0$; Table~\ref{tab:BID}). The human check is one author on
50 Qwen $T=1.0$ answer pairs at the adherence 3 versus 2 boundary
(77.6\% agreement, 95\% interval 63.4\% to 88.2\%), with no second
rater and no check on Llama or multi-turn answers. Judges from other
model families and a larger human evaluation would address these gaps.

\paragraph{Success definition.}
Success requires adherence 3 under the distinctiveness gate,
coherence at least 75 and no degenerate text, and we did not test
sensitivity to the gate or a looser threshold. Below-threshold movement
is visible in the full score distribution for one temperature per
model (Table~\ref{tab:X2}) and in the exploratory sweep, where Llama's
mean adherence exceeds the unsteered level in 0 of 103 site-strength
combinations (Table~\ref{tab:I2c}), but was not examined for Llama at
$T=0.5$ and $1.0$.

\paragraph{Llama.}
Llama's vectors steer only under greedy decoding ($+7.0$~pp at
$T=0.0$, 540 answer pairs). We rule out the implementation, a
role-specific failure and the vector geometry, but do not establish
the cause; the French control's criterion was amended after its output
was inspected (all readings in Table~\ref{tab:I2b}). Since
\citet{chen2025} steer the same model with a two-sided contrast and a
continuous score, the weakness may be specific to our setup; other
sites, lower doses, a two-sided contrast and injection on prompt tokens
could test this. There is no Llama $T=1.5$ setting.

\paragraph{Strengths.}
The chosen prompt-plus-vector strength, 0.15, was the lowest grid
value wherever a strength passed the coherence floor except Llama
$T=0.0$ (0.25), and it was also the lowest over 30 turns for every
Llama condition and Qwen prompt plus vector, so lower doses are
untested; no strength passed the floor at Qwen $T=1.5$ or Llama
$T=1.0$. One strength per setting serves all six roles, although the
role-specific effect ranges from $+3.3$ to $+50.0$~pp across roles
(Qwen, $T=1.0$).

\paragraph{Site and dose.}
On Qwen the injection site lies outside the layers selected at extraction, and the calibration evidence for that choice is suggestive only (Appendix~\ref{appendix}). The activation-level tests were run at the site after it was chosen on steering results, and the site sweep and dose analysis are exploratory.

\paragraph{Multi-turn.}
Each condition has 18 conversations of 30 turns at one temperature per
model, judged on the answer's own rubric only, so the comparisons have
limited power. The level windows were chosen after the first Qwen
results, so the late-turn advantage of steering alone on Qwen
($+13.3$~pp over turns 26 to 30, $p=0.044$) is suggestive even though
it holds in every check in Table~\ref{tab:H3}, and the rounding before
the Wilcoxon test was adopted after the first $p$-values. Cut-off turns
count as failures and hit prompted turns more (35.0\% vs 8.9\% of
turns 21 to 30, Qwen; Table~\ref{tab:H2b}); excluding them changes only
one comparison, kept history versus prompted on Qwen over turns 26 to
30 ($p=0.094$, Appendix~\ref{app:mt-robust}).

\paragraph{Self-naming.}
Llama names its own role outright in roughly a quarter of its
role-prompted answers, inflating the prompt's apparent lead over steering
in some comparisons; excluding these answers does not change the main
single-turn results or the assignment of any extraction to its own role, although rebuilding without them moves the M\&A lawyer and athlete centroids (Table~\ref{tab:SNX}); the full analysis appears in Appendix~\ref{app:selfnaming}.
Multi-turn answers were not checked.

\paragraph{Shared direction.}
Most of each vector lies on a direction shared by all six roles
(Figure~\ref{fig:J1}), and we rule out only answer length and the
format instruction as its content; comparing it with the Assistant
direction of \citet{lu2026} could identify it. The component conditions
were run on Qwen at $T=1.0$ only, with the two parts injected at
unequal sizes.

\paragraph{Vector construction.}
No in-domain-only vector was injected, so the dilution check is
geometric only. On Qwen the role prompt replaces the native identity
line that the default side keeps, so the vector may partly encode its
absence. The questions were culled using unsteered Qwen, which may suit
Qwen better than Llama. At $T=0.0$ the 30 extractions of a role are
near-copies (pairwise cosine 0.995 or above), so the evidence that
independent extractions agree rests on $T>0$.

\paragraph{Statistics.}
No multiple-comparison correction is applied, and single-turn intervals are percentile bootstrap intervals, which can be less accurate than bias-corrected intervals when the resampled differences are skewed.

\section*{Broader Impact Statement}
Role vectors cut both ways. Validated vectors could help developers and auditors check what role a system prompt actually assigns, and whether it persists, from inside the model rather than from outputs alone. The same property makes a vector a channel that prompt inspection cannot see: a role injected through the activations leaves no trace in the prompt. Injection requires access to the model's activations, so this risk concerns parties who run open-weight models rather than users of a hosted application. We tested six benign occupational roles only and release vectors for these roles alone. Checking deployed models at the activation level, not only at the prompt, is one way to make such a channel visible.

\ifarxiv
\section*{Acknowledgments}
This work was supported in part by the Spark Fund of the Center for Research Innovation at
Northeastern University and by Northeastern University's Khoury College of Computer Sciences.
This work was completed in part using the Explorer Cluster, supported by Northeastern
University's Research Computing team.
\fi

\bibliography{custom}
\bibliographystyle{tmlr}

\appendix
\section{Method details}
\label{appendix}

\paragraph{Generation.}
Extraction answers were generated with vLLM 0.19.1 \citep{kwon2023vllm} and injection answers with Transformers 5.17.0 \citep{wolf2020transformers}, in bfloat16 on H200 GPUs; injection reseeds 42 for every
batch of 8 questions, so conditions share a random stream. Llama's chat template
adds a date header to every system message. There is no Llama $T=1.5$ setting:
under the rule in use when the temperature grid was fixed (adherence 3 only,
earlier single-call judge), no role reached 10 qualifying answers in any of 30
runs; under the current rule (adherence 2 or 3), two roles would reach 10 in
3 and 5 of 30 runs (an upper bound). The calibration and evaluation sets were
culled from 42 and 102 candidates using unsteered Qwen.

\paragraph{Extraction.}
The averaged span includes the end-of-turn token. A run needs at least 10 kept
role answers; all 30 runs qualified in every role and setting. The default prompt is the model's native system text plus the closing sentence shared by every role prompt (``Answer concisely in 4-6 bullets or short paragraphs''); Llama has no native system text, so its default prompt is the sentence alone (full prompts in \thesuppplain). Qwen's prompted condition lacks the native identity line that its default side carries.

\paragraph{Steering.}
Reference norms were measured on 42 candidate calibration questions and pooled
over four temperatures: 102.1, 120.8 and 149.7 at Qwen layers 19 to 21, and
8.21 at Llama layer 12. On Qwen at $T = 1.0$, the chosen injection site (layers 19 to 21) lies outside the layers selected at extraction; on 180 paired calibration answers, success at the chosen site minus success at the extraction-selected layers is $+6.1$~pp [$-1.1$, $+13.3$], an interval that includes zero. Coherence floors are 90.78, 90.67, 88.69 and 79.74 for
Qwen at $T=0.0$, $0.5$, $1.0$ and $1.5$, and 88.89, 89.72 and 84.57 for Llama
at $T=0.0$, $0.5$ and $1.0$. Random directions have one Gaussian row per
layer, so their per-layer increments differ from the role vector's. An
other-role condition reproduces the source role's steered answers and is not
reported. The kept-history condition checks each turn's cached tokens against the
history; all 540 turns of both kept-history conditions passed.

\paragraph{Judging.}
Replies are parsed as the first integer in range; no score was parsed from
words. Over all 43 main-condition rows the three trials disagreed on 3.0\% to 8.9\%
of all-rubric rows (Qwen prompt plus vector, judged on all six rubrics: 7.3\%,
6.8\% and 7.5\% at $T=0.0$, $0.5$ and $1.0$) and 1.9\% to 10.0\% of own-rubric
rows. Human-check pairs come from seven Qwen $T=1.0$ conditions, 25 in and 25 out of
domain (seed 20261004); the rater agreed with the judge on 20 of 25 in-domain
and 18 of 24 out-of-domain decisive pairs. Adherence and coherence are scored
at judge temperature 1.0 and combined by the median of three trials; role
identification uses one call per answer at temperature 0.

\paragraph{Metrics and statistics.}
Split-half splits use seed 20261002 plus the role index; the nearest-role
margin interval uses 1,000 bootstrap resamples of runs. Level comparisons drop
turns with a missing score (none occur) and are repeated without turns cut off
at 512 tokens. The single-turn interval is read at 0-based positions 250 and
9,749 of the 10,000 ascending-sorted values (the 251st and 9,750th smallest),
without interpolation. Behavioral and multi-turn statistics were computed under Python 3.11 with
scipy 1.17.1 and reproduce under numpy 2.2.6 and 2.4.6 (largest difference
$1.6\times10^{-17}$); extraction-side analyses and the table and figure
scripts ran under Python 3.12 with numpy 2.2.6 and scipy 1.18.1. 

\paragraph{Released artifacts.}\label{app:artifacts} The code, role vectors, question set, role and default prompts, rubrics, judge prompts, calibration procedure and judged single-turn and multi-turn answers
\ifarxiv
are available at \url{https://doi.org/10.5281/zenodo.23992923},
\else
accompany this submission in anonymized form,
\fi
with the data behind every table; extraction answers, calibration answers and per-answer activations are too large to include and are available on request. The code builds on and modifies the Assistant Axis repository \citep{lu2026}, which its authors release under the MIT license.

\section{Results details}
\label{app:results-details}

\subsection{Out-of-domain scoring}
\label{app:ood-scoring}
Out of domain, most answers of every condition get adherence 2, the middle
grade (Qwen, $T=1.0$: 61--92\% by condition; Table~\ref{tab:X2}). Three
results indicate that the judge still reads role expression on these answers.
First, the rubric still separates conditions out of domain: the steering gain
is $+20.0$~pp, prompt plus vector beats the prompt by $+9.3$~pp, and
role specificity on out-of-domain answers alone is $+22.0$~pp for steering and
$+17.2$~pp for prompting (Qwen, $T=1.0$; Llama, $T=0.0$: prompted $+20.6$,
prompt plus vector $+31.9$; Tables~\ref{tab:S1q} and \ref{tab:S1l}). Second,
role identification, which asks the judge to pick the role rather than score
adherence to a given one, shows the same drop for prompted answers: 94.4\%
identified in domain, 26.4\% out of domain (Qwen, $T=1.0$;
Table~\ref{tab:BID}). Third, a human reader agreed with the judge about as
often out of domain (18 of 24 pairs) as in domain (20 of 25; Table~\ref{tab:M1}). None of these is
fully independent of the judge: the same model scores adherence and
identifies the role, with the rubrics in view, and the human check covers
adherence only, on 50 Qwen answer pairs at $T=1.0$ that the judge had already
scored 3 and 2.

\subsection{Stability at other layers}
\label{app:layers}
\begin{figure}[tp]
\centering
\resultsfig{\figcolwidth}{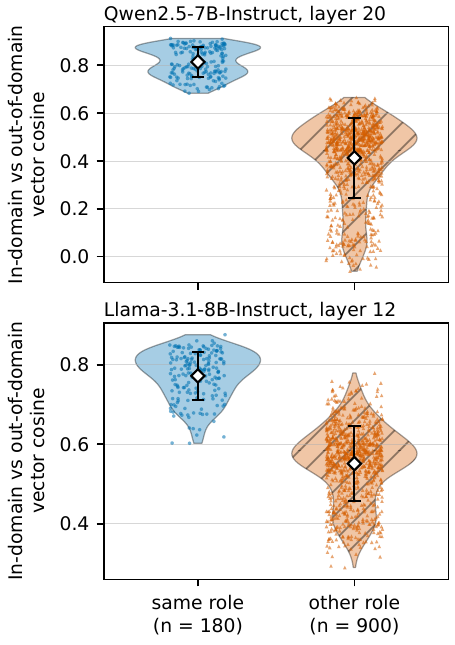}
\caption{Agreement between the in-domain vector (15 questions) and the
out-of-domain vector (75 questions) at the injection site, role vector,
$T=1.0$, 30 extractions per role. Each point is the cosine between the two
vectors of the same extraction: for the same role (180 points, 6 roles
$\times$ 30 extractions) or for a different role (900 points, the in-domain
vector of one role against the out-of-domain vector of each of the other five,
same extraction). Violins show the distributions, white diamonds the mean,
bars one standard deviation. Qwen at layer 20 (top) and Llama at layer 12
(bottom). Means: Qwen same role 0.813, other role 0.413; Llama same role
0.772, other role 0.551.}
\label{fig:D2}
\end{figure}

Table~\ref{tab:A1b} repeats the stability measures of Table~\ref{tab:A1}
at the pilot reference layers (Qwen 18, Llama 21) and at the most
separable band (Qwen 22 to 24, Llama 23 to 27), for the role vector and
the role-specific component; the role-vector values are close to those
at the injection site in every setting. Tables~\ref{tab:B1},
\ref{tab:D1}, \ref{tab:D1b} and \ref{tab:C1} and Figure~\ref{fig:D2}
give the separability, cross-domain and dilution measures at the
injection site in full.

\subsection{Prompt plus vector: domain, length and identification checks}
\label{app:ppv}
The gain comes from out-of-domain questions: on Qwen at
$T=1.0$, in-domain success is 80.0\% in both conditions, while out-of-domain
success rises from 25.6\% to 34.9\% ($+9.3$~pp). Out of domain, role
specificity is higher with prompt plus vector than with the prompt alone in
all five settings, by $+5.9$ to $+11.6$~pp, while in domain, where the
prompt alone is already highly specific, the two are not distinguishable
(Table~\ref{tab:T1}). Prompt plus vector answers are longer (Qwen, $T=1.0$:
256 vs 227 generated tokens). If longer answers simply scored higher,
success would rise on all six rubrics alike; the rise in role specificity
shows the gain is not such a general length effect. On pairs where the
vector answer is no longer than the prompted one, the out-of-domain success
gain remains on Qwen and on Llama at $T=0.5$ but not at $T=0.0$, so on
Llama it is partly driven by longer answers (Table~\ref{tab:G1}). Role
identification rises from 37.8\% to 49.8\% on Qwen at $T=1.0$
($+12.0$~pp) but gives Llama's gain at most weak support ($T=0.0$:
$+2.7$~pp out of domain, not distinguishable; Table~\ref{tab:BID}). The
cost is coherence: on Qwen at $T=1.0$, mean coherence falls from 95.3 to
91.7 and the share of answers below the 75-point threshold rises from
3.1\% to 9.8\%.

\subsection{Multi-turn robustness}
\label{app:mt-robust}
Turns cut off at the 512-token length limit count as failures and are far
more common in late prompted turns (35.0\% vs 8.9\% of turns 21 to 30),
which favors steering, but the late-turn lead of steering alone on Qwen
remains without them ($+14.6$~pp), without the general professor, whose
rubric is the easiest to satisfy, and on the 12 conversations not used in
an earlier calibration (Table~\ref{tab:H3}). Level differences for every
condition are in Table~\ref{tab:H2b}. Excluding cut-off turns changes only one comparison, kept history versus
prompted on Qwen over turns 26 to 30 ($p=0.094$).

\subsection{Exploratory site and strength sweep}
\label{app:sweep}
\begin{figure*}[tp]
\centering
\resultsfig{\textwidth}{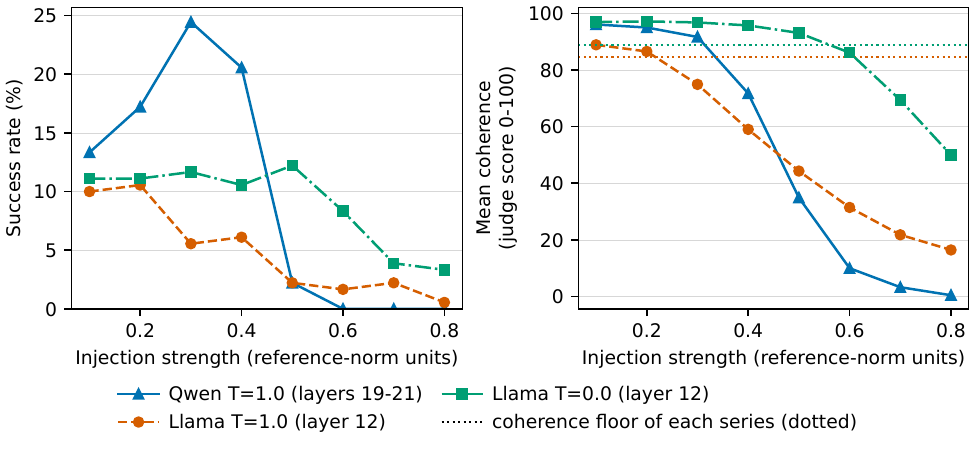}
\caption{Injection-site selection: success rate and mean coherence on the
calibration questions by injection strength at the candidate layers, both
models.}
\label{fig:I1}
\end{figure*}

Across all injection sites and strengths tried, mean adherence never
exceeds the unsteered level on Llama (0 of 103 combinations), against 37
of 70 on Qwen (Table~\ref{tab:I2c}). Figure~\ref{fig:I1} shows success
rate and mean coherence by strength at the candidate layers.

\subsection{French positive control}
\label{app:french}
To test whether injection can steer a simple behavior, we build a
language vector from the 30 calibration questions at $T=1.0$: the mean
activation of answers under the system prompt ``Always answer in
French.'' minus that under the default prompt, averaged over generated
tokens, with no weighting or runs. It is injected like a role vector, at
the same sites and in the same strength units (plus Llama layer 15), at
strengths 0.025 to 0.8. An answer is clean, coherent French if at least
98\% of its letters are Latin script, its French function-word share
exceeds 0.12 and its English share, and its coherence is at least 75; the
control passes if that share exceeds the unsteered share by at least 50
percentage points at some strength. This amended rule is our primary
reading; Table~\ref{tab:I2b} gives all readings. The detector counts
function words over the whole answer, so only the coherence condition
rejects mixed-language answers.
A French vector built and injected the same way yields clean, coherent
French in 76.7\% of Qwen answers (strength 0.2) but in none on Llama at
any strength, where it shifts the language but destroys coherence
(Table~\ref{tab:I2b}).

\subsection{Shared direction and role-specific component}
\label{app:shared}
\begin{figure*}[tp]
\centering
\resultsfig{0.48\textwidth}{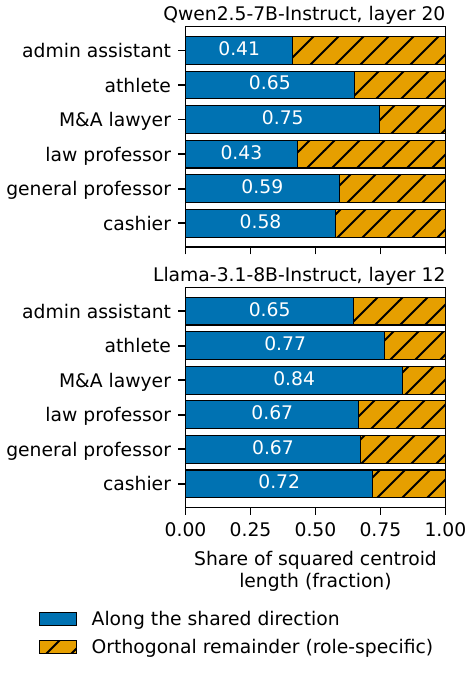}
\hfill
\resultsfig{0.48\textwidth}{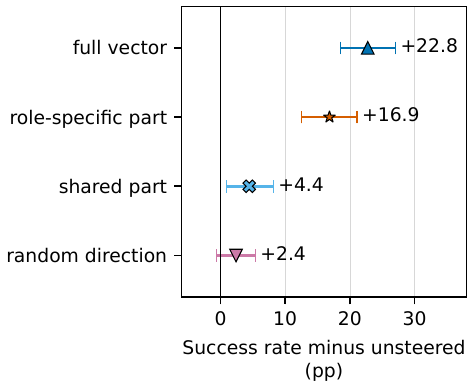}
\caption{What the role vector contains and what steers. Left: share of each
role centroid's squared length along the shared direction (blue) and in the
role-specific component (orange), $T=1.0$, at the injection site. Qwen is
drawn at layer 20, the middle of its injection site (layers 19 to 21); Llama
at its injection layer 12. Appendix Figure~\ref{fig:J1b} draws all three Qwen layers. Right: steering gain (success rate minus unsteered, pp,
95\% intervals), Qwen, $T=1.0$, for the whole vector, its two parts and the
random-direction control.}
\label{fig:J1}
\end{figure*}

The role vector is small: activations under a role prompt are close to
those under the default prompt (cosine 0.98--0.99, both models), and
their difference is about a tenth as long as a typical token's activation.
A large share of it lies on a \emph{shared direction}, the average of the
six role vectors: 36 to 76\% of each vector's squared length on Qwen and
65 to 84\% on Llama at $T=1.0$ (Figure~\ref{fig:J1}, left). The shared
direction is the same at every temperature (cosine 0.97--1.00) and only
weakly related to answer length (within-role $r$ 0.10 to 0.20); it is
consistent with a general shift that accompanies adopting any role,
though we do not identify it further. We injected each part separately on
Qwen at $T=1.0$, its setting with the largest steering gain. The steering
gain is $+22.8$~pp for the whole vector, $+16.9$~pp for the role-specific
part, $+4.4$~pp for the shared part and $+2.4$~pp for the
random-direction control (Figure~\ref{fig:J1}, right). The shared part
steers little for every role (at most $+8.9$~pp), although it was
injected on average about 14\% larger than the role-specific part, whose
gain varies widely by role ($+3.3$ to $+50.0$~pp). The shared part is
thus a large share of the vector's length but a small share of its
effect, and the whole vector steers most.

\subsection{Self-naming}
\label{app:selfnaming}
Llama names its role outright in about a quarter of its role-prompted
answers (Qwen almost never), both in the single-turn conditions and in the
extraction answers that build its vectors; detection uses a text pattern
and is approximate (Table~\ref{tab:SN}).

\paragraph{Steering alone versus the prompt.}
130 of 540 prompted Llama answers (24.1\%) name their own role outright
(e.g.\ ``As a corporate M\&A lawyer''), which no steered answer and almost
no Qwen answer does (Table~\ref{tab:SN}). On Llama at $T=0.0$, on the 331
of 450 out-of-domain questions where neither the steered nor the prompted
answer names the role, the steered minus prompted identification rate is
$-10.3$~pp instead of $-31.6$~pp, so the lead does not rest on
self-naming alone.

\paragraph{Prompt plus vector.}
Self-naming occurs equally often in both Llama conditions at $T=0.0$
(about 24\% of answers; Table~\ref{tab:SN}), and the gain remains when it
is set aside: of the 450 out-of-domain questions, on the 327 where neither
the prompted nor the prompt plus vector answer names the role, the success
gain is $+6.7$~pp [2.2, 11.0] (against $+10.0$~pp on all 450) and the gain
in role specificity $+7.4$~pp [2.8, 12.1] (against $+11.3$~pp).

\paragraph{Extraction.}
For two Llama roles, part of the centroid reflects self-naming answers.
Rebuilding Llama's vectors without them moves some centroids, most for the
M\&A lawyer and the athlete (cosine to the published centroid as low as
0.945; Table~\ref{tab:SNX}), although every extraction is still assigned
to its own role and the designed pairs separate slightly more. The
rebuild removes whole answers, so it does not isolate a self-description
component.

\begin{table*}[tp]
\centering
{\footnotesize\setlength{\tabcolsep}{3pt}
\begin{tabular}{llllr}
\toprule
Model & \shortstack[l]{Condition\\ (strength)} & \shortstack[l]{Level difference,\\ turns 1-30} & \shortstack[l]{Level difference,\\ turns 26-30} & \shortstack[l]{Length-stop\\ share, turns 21-30} \\
\midrule
\multicolumn{5}{l}{\textit{Qwen2.5-7B-Instruct, T=1.0 (18 conversations per condition)}} \\
Qwen & steered (completion, 0.25) & +7.4 [$-$1.3, +16.7], p=0.087 & +13.3 [+2.2, +24.4], p=0.044 & 8.9 \\
Qwen & prompt plus vector (0.15) & +7.8 [+1.9, +14.6], p=0.031 & +2.2 [$-$7.8, +12.2], p=0.719 & 46.1 \\
Qwen & kept history (0.25) & +6.1 [$-$3.9, +16.1], p=0.137 & +7.8 [$-$5.6, +22.2], p=0.365 & 23.3 \\
Qwen & prompted (reference) & reference & reference & 35.0 \\
\midrule
\multicolumn{5}{l}{\textit{Llama-3.1-8B-Instruct, T=0.0 (18 conversations per condition)}} \\
Llama & steered (completion, 0.15) & $-$8.5 [$-$18.7, $-$0.2], p=0.052 & $-$4.4 [$-$21.1, +10.0], p=0.593 & 3.3 \\
Llama & prompt plus vector (0.15) & +4.6 [$-$1.5, +10.7], p=0.107 & +11.1 [+2.2, +21.1], p=0.047 & 23.9 \\
Llama & kept history (0.15) & $-$9.3 [$-$18.7, $-$0.4], p=0.009 & $-$6.7 [$-$23.3, +8.9], p=0.465 & 3.3 \\
Llama & prompted (reference) & reference & reference & 21.1 \\
\bottomrule
\end{tabular}}

\caption{Persistence level against the prompted condition (companion to Table~\ref{tab:H2}), 18 conversations per condition, Qwen $T=1.0$ and
Llama $T=0.0$: the per-conversation mean success rate, condition minus
prompted (pp, 95\% conversation bootstrap interval, tie-robust Wilcoxon
signed-rank $p$), over turns 1 to 30 and 26 to 30, and the share of turns
21 to 30 that stopped at the 512-token limit (percent of turns); a turn
that stops at the limit counts as a failure.}
\label{tab:H2b}
\end{table*}

\begin{table}[tp]
\centering
{\footnotesize\setlength{\tabcolsep}{3pt}
\begin{tabular}{lrrc}
\toprule
\multicolumn{4}{l}{\textbf{Panel A: In-domain ($n{=}90$)}} \\
Model & $T$ & Diff & 95\% CI \\
\midrule
Qwen & 0.0 & +8.0 & [$-$1.3, +17.3] \\
Qwen & 0.5 & +0.9 & [$-$8.7, +10.4] \\
Qwen & 1.0 & +4.7 & [$-$3.8, +13.1] \\
Llama & 0.0 & $-$2.0 & [$-$10.4, +6.4] \\
Llama & 0.5 & $-$6.4 & [$-$14.9, +1.8] \\
\bottomrule
\end{tabular}}
\par\smallskip
{\footnotesize\setlength{\tabcolsep}{3pt}
\begin{tabular}{lrrc}
\toprule
\multicolumn{4}{l}{\textbf{Panel B: Out-of-domain ($n{=}450$)}} \\
Model & $T$ & Diff & 95\% CI \\
\midrule
Qwen & 0.0 & +5.9 & [+1.2, +10.4] \\
Qwen & 0.5 & +8.5 & [+3.9, +13.2] \\
Qwen & 1.0 & +11.6 & [+7.1, +16.4] \\
Llama & 0.0 & +11.3 & [+7.7, +15.0] \\
Llama & 0.5 & +7.9 & [+3.7, +12.1] \\
\bottomrule
\end{tabular}}

\caption{Prompt plus vector minus prompted role specificity (pp, 95\%
question-cluster intervals in brackets) by question domain, for every setting
where prompt plus vector was run. Panel A: in-domain questions (90 answers).
Panel B: out-of-domain questions (450 answers). No multiplicity correction.}
\label{tab:T1}
\end{table}

\begin{table*}[tp]
\centering
\renewcommand{\arraystretch}{0.95}   %
{\footnotesize\setlength{\tabcolsep}{3pt}
\begin{tabular}{lllrrr}
\toprule
\multicolumn{6}{l}{\textbf{Panel A}} \\
Layer set & Model & \shortstack[l]{T\\ (temperature)} & Layer (index) & \shortstack[l]{Pairwise\\ cosine among\\ 30 run\\ vectors, min\\ to max over\\ roles (cosine)} & \shortstack[l]{Split-half\\ centroid\\ cosine, min\\ over roles\\ (cosine)} \\
\midrule
\multicolumn{6}{l}{\textit{Role vector}} \\
injection site & Qwen2.5-7B-Instruct & 0.0 & 19, 20, 21 & 0.997 to 1.000 & 0.9998 \\
reference layer & Qwen2.5-7B-Instruct & 0.0 & 18 & 0.997 to 1.000 & 0.9998 \\
most separable band & Qwen2.5-7B-Instruct & 0.0 & 22, 23, 24 & 0.997 to 1.000 & 0.9998 \\
\addlinespace
injection site & Qwen2.5-7B-Instruct & 0.5 & 19, 20, 21 & 0.938 to 0.984 & 0.9957 \\
reference layer & Qwen2.5-7B-Instruct & 0.5 & 18 & 0.936 to 0.977 & 0.9955 \\
most separable band & Qwen2.5-7B-Instruct & 0.5 & 22, 23, 24 & 0.929 to 0.985 & 0.9950 \\
\addlinespace
injection site & Qwen2.5-7B-Instruct & 1.0 & 19, 20, 21 & 0.915 to 0.977 & 0.9939 \\
reference layer & Qwen2.5-7B-Instruct & 1.0 & 18 & 0.909 to 0.967 & 0.9934 \\
most separable band & Qwen2.5-7B-Instruct & 1.0 & 22, 23, 24 & 0.904 to 0.980 & 0.9931 \\
\addlinespace
injection site & Qwen2.5-7B-Instruct & 1.5 & 19, 20, 21 & 0.898 to 0.964 & 0.9927 \\
reference layer & Qwen2.5-7B-Instruct & 1.5 & 18 & 0.891 to 0.952 & 0.9922 \\
most separable band & Qwen2.5-7B-Instruct & 1.5 & 22, 23, 24 & 0.885 to 0.966 & 0.9917 \\
\addlinespace
injection site & Llama-3.1-8B-Instruct & 0.0 & 12 & 0.995 to 0.998 & 0.9997 \\
reference layer & Llama-3.1-8B-Instruct & 0.0 & 21 & 0.994 to 0.999 & 0.9996 \\
most separable band & Llama-3.1-8B-Instruct & 0.0 & 23, 24, 25, 26, 27 & 0.993 to 0.999 & 0.9996 \\
\addlinespace
injection site & Llama-3.1-8B-Instruct & 0.5 & 12 & 0.930 to 0.969 & 0.9950 \\
reference layer & Llama-3.1-8B-Instruct & 0.5 & 21 & 0.919 to 0.978 & 0.9941 \\
most separable band & Llama-3.1-8B-Instruct & 0.5 & 23, 24, 25, 26, 27 & 0.911 to 0.981 & 0.9934 \\
\addlinespace
injection site & Llama-3.1-8B-Instruct & 1.0 & 12 & 0.907 to 0.954 & 0.9932 \\
reference layer & Llama-3.1-8B-Instruct & 1.0 & 21 & 0.892 to 0.963 & 0.9920 \\
most separable band & Llama-3.1-8B-Instruct & 1.0 & 23, 24, 25, 26, 27 & 0.880 to 0.967 & 0.9910 \\
\midrule
\multicolumn{6}{l}{\textit{Role-specific vector (shared cross-role direction removed)}} \\
injection site & Qwen2.5-7B-Instruct & 0.0 & 19, 20, 21 & 0.990 to 0.999 & 0.9993 \\
reference layer & Qwen2.5-7B-Instruct & 0.0 & 18 & 0.984 to 0.999 & 0.9989 \\
most separable band & Qwen2.5-7B-Instruct & 0.0 & 22, 23, 24 & 0.992 to 0.999 & 0.9995 \\
\addlinespace
injection site & Qwen2.5-7B-Instruct & 0.5 & 19, 20, 21 & 0.903 to 0.971 & 0.9930 \\
reference layer & Qwen2.5-7B-Instruct & 0.5 & 18 & 0.844 to 0.952 & 0.9879 \\
most separable band & Qwen2.5-7B-Instruct & 0.5 & 22, 23, 24 & 0.908 to 0.976 & 0.9934 \\
\addlinespace
injection site & Qwen2.5-7B-Instruct & 1.0 & 19, 20, 21 & 0.872 to 0.960 & 0.9908 \\
reference layer & Qwen2.5-7B-Instruct & 1.0 & 18 & 0.792 to 0.932 & 0.9825 \\
most separable band & Qwen2.5-7B-Instruct & 1.0 & 22, 23, 24 & 0.880 to 0.968 & 0.9914 \\
\addlinespace
injection site & Qwen2.5-7B-Instruct & 1.5 & 19, 20, 21 & 0.839 to 0.937 & 0.9876 \\
reference layer & Qwen2.5-7B-Instruct & 1.5 & 18 & 0.727 to 0.892 & 0.9756 \\
most separable band & Qwen2.5-7B-Instruct & 1.5 & 22, 23, 24 & 0.849 to 0.946 & 0.9887 \\
\addlinespace
injection site & Llama-3.1-8B-Instruct & 0.0 & 12 & 0.986 to 0.995 & 0.9991 \\
reference layer & Llama-3.1-8B-Instruct & 0.0 & 21 & 0.991 to 0.998 & 0.9994 \\
most separable band & Llama-3.1-8B-Instruct & 0.0 & 23, 24, 25, 26, 27 & 0.992 to 0.998 & 0.9994 \\
\addlinespace
injection site & Llama-3.1-8B-Instruct & 0.5 & 12 & 0.844 to 0.908 & 0.9879 \\
reference layer & Llama-3.1-8B-Instruct & 0.5 & 21 & 0.891 to 0.960 & 0.9918 \\
most separable band & Llama-3.1-8B-Instruct & 0.5 & 23, 24, 25, 26, 27 & 0.899 to 0.967 & 0.9925 \\
\addlinespace
injection site & Llama-3.1-8B-Instruct & 1.0 & 12 & 0.769 to 0.863 & 0.9801 \\
reference layer & Llama-3.1-8B-Instruct & 1.0 & 21 & 0.864 to 0.931 & 0.9896 \\
most separable band & Llama-3.1-8B-Instruct & 1.0 & 23, 24, 25, 26, 27 & 0.873 to 0.943 & 0.9904 \\
\bottomrule
\end{tabular}}

\caption{Stability at the injection site compared with the pilot reference
layers (Qwen 18, Llama 21) and the most separable band (Qwen 22 to 24, Llama
23 to 27), per sampling temperature, 30 extractions per role and six roles,
for the role vector and the role-specific component. Columns as in
Table~\ref{tab:A1}, printed as two panels: Panel A (this table) the pairwise
and split-half cosines, Panel D (Table~\ref{tab:A1b} continued) the lowest
role and the cosine to the $T=0.0$ centroid. For a layer set with more than
one layer, each column is aggregated over that set separately (pairwise-cosine
range from the lowest minimum to the highest maximum, every other column the
worst value), so one row may mix layers. Leave-one-out assignment is 100\% in
every setting, so its two panels (B and C) are not printed.}
\label{tab:A1b}
\end{table*}

\begin{table*}[tp]
\centering
{\footnotesize\setlength{\tabcolsep}{3pt}
\begin{tabular}{lllrlr}
\toprule
\multicolumn{6}{l}{\textbf{Panel D}} \\
Layer set & Model & \shortstack[l]{T\\ (temperature)} & Layer (index) & Lowest role (name) & \shortstack[l]{Cosine to\\ T=0.0\\ centroid,\\ min over\\ roles\\ (cosine)} \\
\midrule
\multicolumn{6}{l}{\textit{Role vector}} \\
injection site & Qwen2.5-7B-Instruct & 0.0 & 19, 20, 21 & none (all roles 100\%) & 1 (reference) \\
reference layer & Qwen2.5-7B-Instruct & 0.0 & 18 & none (all roles 100\%) & 1 (reference) \\
most separable band & Qwen2.5-7B-Instruct & 0.0 & 22, 23, 24 & none (all roles 100\%) & 1 (reference) \\
\addlinespace
injection site & Qwen2.5-7B-Instruct & 0.5 & 19, 20, 21 & none (all roles 100\%) & 0.976 \\
reference layer & Qwen2.5-7B-Instruct & 0.5 & 18 & none (all roles 100\%) & 0.976 \\
most separable band & Qwen2.5-7B-Instruct & 0.5 & 22, 23, 24 & none (all roles 100\%) & 0.972 \\
\addlinespace
injection site & Qwen2.5-7B-Instruct & 1.0 & 19, 20, 21 & none (all roles 100\%) & 0.971 \\
reference layer & Qwen2.5-7B-Instruct & 1.0 & 18 & none (all roles 100\%) & 0.975 \\
most separable band & Qwen2.5-7B-Instruct & 1.0 & 22, 23, 24 & none (all roles 100\%) & 0.963 \\
\addlinespace
injection site & Qwen2.5-7B-Instruct & 1.5 & 19, 20, 21 & none (all roles 100\%) & 0.954 \\
reference layer & Qwen2.5-7B-Instruct & 1.5 & 18 & none (all roles 100\%) & 0.942 \\
most separable band & Qwen2.5-7B-Instruct & 1.5 & 22, 23, 24 & none (all roles 100\%) & 0.944 \\
\addlinespace
injection site & Llama-3.1-8B-Instruct & 0.0 & 12 & none (all roles 100\%) & 1 (reference) \\
reference layer & Llama-3.1-8B-Instruct & 0.0 & 21 & none (all roles 100\%) & 1 (reference) \\
most separable band & Llama-3.1-8B-Instruct & 0.0 & 23, 24, 25, 26, 27 & none (all roles 100\%) & 1 (reference) \\
\addlinespace
injection site & Llama-3.1-8B-Instruct & 0.5 & 12 & none (all roles 100\%) & 0.971 \\
reference layer & Llama-3.1-8B-Instruct & 0.5 & 21 & none (all roles 100\%) & 0.969 \\
most separable band & Llama-3.1-8B-Instruct & 0.5 & 23, 24, 25, 26, 27 & none (all roles 100\%) & 0.966 \\
\addlinespace
injection site & Llama-3.1-8B-Instruct & 1.0 & 12 & none (all roles 100\%) & 0.950 \\
reference layer & Llama-3.1-8B-Instruct & 1.0 & 21 & none (all roles 100\%) & 0.949 \\
most separable band & Llama-3.1-8B-Instruct & 1.0 & 23, 24, 25, 26, 27 & none (all roles 100\%) & 0.944 \\
\midrule
\multicolumn{6}{l}{\textit{Role-specific vector (shared cross-role direction removed)}} \\
injection site & Qwen2.5-7B-Instruct & 0.0 & 19, 20, 21 & none (all roles 100\%) & 1 (reference) \\
reference layer & Qwen2.5-7B-Instruct & 0.0 & 18 & none (all roles 100\%) & 1 (reference) \\
most separable band & Qwen2.5-7B-Instruct & 0.0 & 22, 23, 24 & none (all roles 100\%) & 1 (reference) \\
\addlinespace
injection site & Qwen2.5-7B-Instruct & 0.5 & 19, 20, 21 & none (all roles 100\%) & 0.976 \\
reference layer & Qwen2.5-7B-Instruct & 0.5 & 18 & none (all roles 100\%) & 0.963 \\
most separable band & Qwen2.5-7B-Instruct & 0.5 & 22, 23, 24 & none (all roles 100\%) & 0.978 \\
\addlinespace
injection site & Qwen2.5-7B-Instruct & 1.0 & 19, 20, 21 & none (all roles 100\%) & 0.959 \\
reference layer & Qwen2.5-7B-Instruct & 1.0 & 18 & none (all roles 100\%) & 0.952 \\
most separable band & Qwen2.5-7B-Instruct & 1.0 & 22, 23, 24 & none (all roles 100\%) & 0.960 \\
\addlinespace
injection site & Qwen2.5-7B-Instruct & 1.5 & 19, 20, 21 & none (all roles 100\%) & 0.929 \\
reference layer & Qwen2.5-7B-Instruct & 1.5 & 18 & none (all roles 100\%) & 0.921 \\
most separable band & Qwen2.5-7B-Instruct & 1.5 & 22, 23, 24 & none (all roles 100\%) & 0.935 \\
\addlinespace
injection site & Llama-3.1-8B-Instruct & 0.0 & 12 & none (all roles 100\%) & 1 (reference) \\
reference layer & Llama-3.1-8B-Instruct & 0.0 & 21 & none (all roles 100\%) & 1 (reference) \\
most separable band & Llama-3.1-8B-Instruct & 0.0 & 23, 24, 25, 26, 27 & none (all roles 100\%) & 1 (reference) \\
\addlinespace
injection site & Llama-3.1-8B-Instruct & 0.5 & 12 & none (all roles 100\%) & 0.944 \\
reference layer & Llama-3.1-8B-Instruct & 0.5 & 21 & none (all roles 100\%) & 0.968 \\
most separable band & Llama-3.1-8B-Instruct & 0.5 & 23, 24, 25, 26, 27 & none (all roles 100\%) & 0.970 \\
\addlinespace
injection site & Llama-3.1-8B-Instruct & 1.0 & 12 & none (all roles 100\%) & 0.922 \\
reference layer & Llama-3.1-8B-Instruct & 1.0 & 21 & none (all roles 100\%) & 0.941 \\
most separable band & Llama-3.1-8B-Instruct & 1.0 & 23, 24, 25, 26, 27 & none (all roles 100\%) & 0.934 \\
\bottomrule
\end{tabular}}

\caption{Table~\ref{tab:A1b} (continued): Panel D, lowest role and cosine to
the $T=0.0$ centroid.}
\end{table*}

\begin{table*}[tp]
\centering
{\footnotesize\setlength{\tabcolsep}{2pt}
\begin{tabular}{lrrrrrr}
\toprule
Model & \shortstack[l]{Layer\\ (index)} & \shortstack[l]{Angle,\\ M\&A lawyer\\ vs law\\ professor\\ (degrees)} & \shortstack[l]{Angle,\\ general\\ professor vs\\ law professor\\ (degrees)} & \shortstack[l]{Margin, M\&A lawyer,\\ 95\% interval\\ (cosine)} & \shortstack[l]{Margin, law professor,\\ 95\% interval\\ (cosine)} & \shortstack[l]{Margin, general professor,\\ 95\% interval\\ (cosine)} \\
\midrule
\multicolumn{7}{l}{\textit{Role vector}} \\
Qwen & 19 & 48.8 & 54.4 & 0.337 [0.329, 0.346] & 0.338 [0.330, 0.347] & 0.381 [0.369, 0.396] \\
Qwen & 20 & 50.5 & 56.8 & 0.358 [0.351, 0.367] & 0.360 [0.353, 0.369] & 0.384 [0.373, 0.398] \\
Qwen & 21 & 53.1 & 58.1 & 0.393 [0.385, 0.403] & 0.395 [0.387, 0.406] & 0.397 [0.384, 0.412] \\
Llama & 12 & 37.1 & 41.9 & 0.198 [0.190, 0.209] & 0.197 [0.189, 0.208] & 0.244 [0.238, 0.256] \\
\midrule
\multicolumn{7}{l}{\textit{Role-specific part (shared cross-role direction removed)}} \\
Qwen & 19 & 77.5 & 84.6 & 0.740 [0.728, 0.758] & 0.768 [0.752, 0.787] & 0.856 [0.840, 0.875] \\
Qwen & 20 & 78.1 & 86.5 & 0.748 [0.737, 0.766] & 0.778 [0.763, 0.797] & 0.884 [0.870, 0.903] \\
Qwen & 21 & 80.2 & 86.7 & 0.782 [0.770, 0.800] & 0.814 [0.798, 0.833] & 0.883 [0.869, 0.903] \\
Llama & 12 & 75.5 & 83.7 & 0.668 [0.647, 0.698] & 0.691 [0.669, 0.721] & 0.652 [0.633, 0.682] \\
\bottomrule
\end{tabular}}

\caption{Designed close pairs at the injection site, $T=1.0$, for the role
vector and the role-specific component (the role vector minus the mean of the
six role centroids at the same model, temperature and layer). Angular
separation between the two roles' centroids (degrees) for M\&A lawyer vs law
professor and for general professor vs law professor, and the nearest-role
margin of each of the three roles involved: the mean over its 30 extractions
of the cosine to its own centroid minus the highest, over the other roles, of
the mean cosine to that role's centroid; 95\% interval from 1,000 bootstrap
resamples of the extractions within each role (for the role-specific component
the mean of the six centroids is recomputed in each resample). Qwen is shown
separately for layers 19, 20 and 21; Llama at layer 12.}
\label{tab:B1}
\end{table*}

\begin{table*}[tp]
\centering
{\footnotesize\setlength{\tabcolsep}{3pt}
\begin{tabular}{llrrrr}
\toprule
Model & T & \shortstack[l]{Same role,\\ in-domain vs\\ out-of-domain\\ vectors: mean\\ $\pm$ SD (cosine)} & \shortstack[l]{Different role,\\ in-domain vs\\ out-of-domain\\ vectors: mean\\ $\pm$ SD (cosine)} & \shortstack[l]{Out-of-domain\\ to in-domain\\ assignment,\\ all roles\\ (\% correct)} & \shortstack[l]{Out-of-domain\\ to in-domain,\\ lowest role:\\ accuracy (name)\\ (\% correct)} \\
\midrule
Qwen & 0.0 & 0.783 $\pm$ 0.069 & 0.360 $\pm$ 0.164 & 100.0 & 100.0 (all roles) \\
Qwen & 0.5 & 0.793 $\pm$ 0.068 & 0.371 $\pm$ 0.166 & 100.0 & 100.0 (all roles) \\
Qwen & 1.0 & 0.791 $\pm$ 0.067 & 0.372 $\pm$ 0.168 & 100.0 & 100.0 (all roles) \\
Qwen & 1.5 & 0.781 $\pm$ 0.071 & 0.377 $\pm$ 0.166 & 100.0 & 100.0 (all roles) \\
\addlinespace
Llama & 0.0 & 0.780 $\pm$ 0.054 & 0.558 $\pm$ 0.104 & 100.0 & 100.0 (all roles) \\
Llama & 0.5 & 0.784 $\pm$ 0.062 & 0.558 $\pm$ 0.105 & 85.0 & 10.0 (cashier) \\
Llama & 1.0 & 0.772 $\pm$ 0.061 & 0.551 $\pm$ 0.094 & 86.7 & 20.0 (cashier) \\
\bottomrule
\end{tabular}}

\caption{Cross-domain consistency of the role vectors at the injection site,
by sampling temperature, 30 extractions per role. Each extraction yields an
in-domain vector (its 15 in-domain questions) and an out-of-domain vector (its
75 out-of-domain questions). Same role: cosine between the two vectors of the
same role in the same extraction (180 pairs); different role: the in-domain
vector of one role against the out-of-domain vector of each other role in the
same extraction (900 pairs); mean and SD. Assignment: each out-of-domain
vector is matched by cosine to the six in-domain role centroids (each the mean
over the 30 extractions, with the vector's own extraction left out of its own
role's centroid) and is correct when the nearest centroid is its own role;
shown for all roles and for the lowest role. Assignment in the other
direction, in-domain vectors to out-of-domain centroids, is 100\% in every
row. Qwen rows aggregate layers 19 to 21 (lowest mean and accuracy, largest
SD), so one row may mix layers; Llama is layer 12. Appendix
Table~\ref{tab:D1b} gives the role-specific component with both assignment
directions.}
\label{tab:D1}
\end{table*}

\begin{table*}[tp]
\centering
{\footnotesize\setlength{\tabcolsep}{3pt}
\textit{Panel A. Llama extraction answers that name their own role in the first person (all 30 extractions, six roles)}\par\smallskip
\begin{tabular}{lrrrr}
\toprule
Temperature & \shortstack[r]{In-domain\\ answers} & \shortstack[r]{Out-of-domain\\ answers} & \shortstack[r]{All\\ answers} & \shortstack[r]{Share of\\ extraction weight} \\
\midrule
0.0 & 451 (16.7\%) & 3,989 (29.7\%) & 4,440 (27.5\%) & 24.2\% \\
0.5 & 427 (15.8\%) & 3,758 (28.0\%) & 4,185 (26.0\%) & 22.8\% \\
1.0 & 321 (12.2\%) & 2,815 (21.7\%) & 3,136 (20.1\%) & 17.6\% \\
\bottomrule
\end{tabular}
\par\bigskip
\textit{Panel B. Llama vectors at layer 12 rebuilt without those answers, compared with the published vectors}\par\smallskip
\begin{tabular}{llrrrrr}
\toprule
Temperature & Variant & \shortstack[r]{Lowest cosine\\ to published\\ centroid (role)} & \shortstack[r]{Leave-one-\\ run-out\\ accuracy} & \shortstack[r]{Same-role\\ pairwise cosine,\\ published} & \shortstack[r]{Same-role\\ pairwise cosine,\\ rebuilt} & \shortstack[r]{Angle change,\\ M\&A lawyer vs\\ law professor} \\
\midrule
0.0 & role vector    & 0.9452 (M\&A lawyer) & 100\% & 0.970 to 1.000 & 0.961 to 1.000 & $+3.56^{\circ}$ \\
0.0 & role-specific  & 0.8508 (M\&A lawyer) & 100\% & 0.923 to 1.000 & 0.850 to 1.000 & $+5.94^{\circ}$ \\
0.5 & role vector    & 0.9743 (athlete)     & 100\% & 0.897 to 0.979 & 0.564 to 0.979 & $+2.67^{\circ}$ \\
0.5 & role-specific  & 0.9352 (athlete)     & 100\% & 0.726 to 0.939 & 0.193 to 0.941 & $+4.98^{\circ}$ \\
1.0 & role vector    & 0.9825 (athlete)     & 100\% & 0.855 to 0.973 & 0.451 to 0.970 & $+2.87^{\circ}$ \\
1.0 & role-specific  & 0.9561 (athlete)     & 100\% & 0.547 to 0.918 & 0.108 to 0.910 & $+5.54^{\circ}$ \\
\bottomrule
\end{tabular}}

\caption{Self-naming in Llama's extraction answers and its effect on the
extracted vectors. Detection uses the same first-person pattern as
Table~\ref{tab:SN} plus the ``as your'' marker (for example ``As your M\&A
lawyer''). Panel A: Llama extraction answers that self-name, over all 30
extractions and six roles, by temperature: count and share of in-domain,
out-of-domain and all extraction answers, and their share of the extraction
weight (each role in each extraction has total weight 1; an answer counts in
proportion to its adherence score, 2 or 3, within its question group; the
in-domain and out-of-domain groups count half each). Panel B: Llama vectors
at layer 12 rebuilt from the stored per-answer activations without the
self-naming answers, compared with the published vectors; the role vector
keeps the shared direction, the role-specific component has it removed, with
the shared direction recomputed for each set. Columns: lowest cosine between
the rebuilt and the published role centroid over the six roles (role named);
lowest leave-one-run-out accuracy; range of same-role pairwise cosines between
extractions, published and rebuilt; change in the angle between the M\&A
lawyer and corporations law professor centroids (positive: further apart
after rebuilding). Three competitive athlete extractions (two at $T=0.5$, one
at $T=1.0$) cannot be rebuilt because every weighted out-of-domain answer in
them self-names; that role uses its remaining 28 or 29 extractions. The low
rebuilt pairwise minimums at $T=0.5$ and $1.0$ come from extractions that keep
only 1 to 5 out-of-domain answers; extractions keeping 60 or more answers have
cosine at least 0.991 to their published extraction. At $T=0.0$ the 30
extractions give identical answers, so the same answers self-name in every
run and the rebuild changes the question mix rather than thinning
extractions. The shared direction's share of squared length, mean over roles,
published vs rebuilt: 0.704 vs 0.701 ($T=0.0$), 0.715 vs 0.720 ($T=0.5$),
0.718 vs 0.720 ($T=1.0$). On Qwen, 0.0\% to 0.1\% of extraction answers
self-name at every temperature; rebuilt centroids have cosine 1.0000 to the
published ones at layers 19 to 21, and the designed-pair angles change by at
most 0.01 degrees.}
\label{tab:SNX}
\end{table*}

\begin{table*}[tp]
\centering
{\footnotesize\setlength{\tabcolsep}{2pt}
\begin{tabular}{llrrrrrrr}
\toprule
Model & T & \shortstack[l]{Layer\\ (index)} & \shortstack[l]{Same role,\\ in-domain vs\\ out-of-domain\\ vectors: mean\\ $\pm$ SD (cosine)} & \shortstack[l]{Different role,\\ in-domain vs\\ out-of-domain\\ vectors: mean\\ $\pm$ SD (cosine)} & \shortstack[l]{In-domain to\\ out-of-domain\\ assignment,\\ all roles\\ (\% correct)} & \shortstack[l]{In-domain to\\ out-of-domain,\\ lowest role:\\ accuracy (name)\\ (\% correct)} & \shortstack[l]{Out-of-domain\\ to in-domain\\ assignment,\\ all roles\\ (\% correct)} & \shortstack[l]{Out-of-domain\\ to in-domain,\\ lowest role:\\ accuracy (name)\\ (\% correct)} \\
\midrule
Qwen & 0.0 & 19 to 21 & 0.701 $\pm$ 0.087 & $-$0.139 $\pm$ 0.269 & 100.0 & 100.0 (all roles) & 100.0 & 100.0 (all roles) \\
Qwen & 0.5 & 19 to 21 & 0.712 $\pm$ 0.083 & $-$0.117 $\pm$ 0.264 & 100.0 & 100.0 (all roles) & 100.0 & 100.0 (all roles) \\
Qwen & 1.0 & 19 to 21 & 0.704 $\pm$ 0.083 & $-$0.107 $\pm$ 0.259 & 100.0 & 100.0 (all roles) & 100.0 & 100.0 (all roles) \\
Qwen & 1.5 & 19 to 21 & 0.680 $\pm$ 0.091 & $-$0.097 $\pm$ 0.251 & 100.0 & 100.0 (all roles) & 100.0 & 100.0 (all roles) \\
\addlinespace
Llama & 0.0 & 12 & 0.546 $\pm$ 0.096 & $-$0.105 $\pm$ 0.208 & 100.0 & 100.0 (all roles) & 100.0 & 100.0 (all roles) \\
Llama & 0.5 & 12 & 0.557 $\pm$ 0.084 & $-$0.082 $\pm$ 0.207 & 100.0 & 100.0 (all roles) & 100.0 & 100.0 (all roles) \\
Llama & 1.0 & 12 & 0.542 $\pm$ 0.083 & $-$0.065 $\pm$ 0.198 & 100.0 & 100.0 (all roles) & 100.0 & 100.0 (all roles) \\
\bottomrule
\end{tabular}}

\caption{Cross-domain consistency of the role-specific component (the role
vector minus the direction shared by all roles) at the injection site, by
sampling temperature, 30 extractions per role; same quantities, layers and
aggregation as Table~\ref{tab:D1}. Both assignment directions are shown: each
out-of-domain (or in-domain) vector is matched by cosine to the six in-domain
(or out-of-domain) role centroids, each the mean over the 30 extractions with
the vector's own extraction left out of its own role's centroid, and is
correct when the nearest centroid is its own role; all roles, and the lowest
role with its name.}
\label{tab:D1b}
\end{table*}

\begin{table*}[tp]
\centering
{\footnotesize\setlength{\tabcolsep}{3pt}
\begin{tabular}{llllr}
\toprule
\multicolumn{5}{l}{\textbf{Panel A}} \\
Model & \shortstack[l]{T\\ (temperature)} & \shortstack[l]{Layer\\ (index)} & Construction & \shortstack[l]{Pairwise\\ cosine\\ among 30\\ run vectors,\\ mean over\\ roles\\ (cosine)} \\
\midrule
Qwen2.5-7B-Instruct & 1.0 & 19 to 21 & in-domain only (15 questions) & 0.893 \\
Qwen2.5-7B-Instruct & 1.0 & 19 to 21 & out-of-domain only, 75 questions & 0.964 \\
Qwen2.5-7B-Instruct & 1.0 & 19 to 21 & \shortstack[l]{balanced, official (15 in-domain\\ + 75 out-of-domain, 0.5/0.5)} & 0.953 \\
Qwen2.5-7B-Instruct & 1.0 & 19 to 21 & \shortstack[l]{out-of-domain only, 15 questions\\ (matched size)} & 0.888 \\
\addlinespace
Llama-3.1-8B-Instruct & 1.0 & 12 & in-domain only (15 questions) & 0.812 \\
Llama-3.1-8B-Instruct & 1.0 & 12 & out-of-domain only, 75 questions & 0.956 \\
Llama-3.1-8B-Instruct & 1.0 & 12 & \shortstack[l]{balanced, official (15 in-domain\\ + 75 out-of-domain, 0.5/0.5)} & 0.933 \\
Llama-3.1-8B-Instruct & 1.0 & 12 & \shortstack[l]{out-of-domain only, 15 questions\\ (matched size)} & 0.838 \\
\bottomrule
\end{tabular}}
\par\smallskip
{\footnotesize\setlength{\tabcolsep}{3pt}
\begin{tabular}{llllrr}
\toprule
\multicolumn{6}{l}{\textbf{Panel B}} \\
Model & \shortstack[l]{T\\ (temperature)} & \shortstack[l]{Layer\\ (index)} & Construction & \shortstack[l]{Leave-one-run-out\\ accuracy, mean\\ (lowest role) (\%\\ correct)} & \shortstack[l]{Angle,\\ M\&A\\ lawyer vs\\ law\\ professor\\ (degrees)} \\
\midrule
Qwen2.5-7B-Instruct & 1.0 & 19 to 21 & in-domain only (15 questions) & 100.0 (100.0) & 49.3 to 53.6 \\
Qwen2.5-7B-Instruct & 1.0 & 19 to 21 & out-of-domain only, 75 questions & 100.0 (100.0) & 49.5 to 54.3 \\
Qwen2.5-7B-Instruct & 1.0 & 19 to 21 & \shortstack[l]{balanced, official (15 in-domain\\ + 75 out-of-domain, 0.5/0.5)} & 100.0 (100.0) & 48.8 to 53.1 \\
Qwen2.5-7B-Instruct & 1.0 & 19 to 21 & \shortstack[l]{out-of-domain only, 15 questions\\ (matched size)} & 100.0 (100.0) & 51.0 to 55.8 \\
\addlinespace
Llama-3.1-8B-Instruct & 1.0 & 12 & in-domain only (15 questions) & 100.0 (100.0) & 40.6 \\
Llama-3.1-8B-Instruct & 1.0 & 12 & out-of-domain only, 75 questions & 100.0 (100.0) & 37.5 \\
Llama-3.1-8B-Instruct & 1.0 & 12 & \shortstack[l]{balanced, official (15 in-domain\\ + 75 out-of-domain, 0.5/0.5)} & 100.0 (100.0) & 37.1 \\
Llama-3.1-8B-Instruct & 1.0 & 12 & \shortstack[l]{out-of-domain only, 15 questions\\ (matched size)} & 100.0 (100.0) & 38.9 \\
\bottomrule
\end{tabular}}

\caption{Dilution check at the injection site, role vector, $T=1.0$, 30
extractions per role. Four versions of the role centroid are built from the
same extractions: in-domain only (15 questions), out-of-domain only (75
questions), the balanced injected centroid (15 in-domain and 75 out-of-domain
questions, weighted 0.5 and 0.5), and out-of-domain only restricted to 15
randomly chosen questions per role (matched size). For each version: the
same-role pairwise cosine (mean over roles), leave-one-out assignment accuracy
(mean, and lowest role in brackets), the angular separation of the two
designed pairs (M\&A lawyer vs law professor; general professor vs law
professor), the lowest 95\% lower bound of the nearest-role margin over the
six roles, and the cosine of the version's centroid to the injected centroid
(mean, and range over roles). Qwen values cover layers 19 to 21 (lowest value
for cosines, accuracies and bounds; range over the layers for angles); Llama
values are for layer 12. Printed as four panels over two floats (A and B here;
C and D in Table~\ref{tab:C1} continued). In the 15-question out-of-domain
control some role and extraction combinations have fewer than 15 qualifying
questions, because answers that failed the quality or adherence rules were
dropped: 50 of 180 combinations on Qwen (fewest 12 questions) and 75 of 180 on
Llama (fewest 7 questions).}
\label{tab:C1}
\end{table*}

\begin{table*}[tp]
\centering
{\footnotesize\setlength{\tabcolsep}{3pt}
\begin{tabular}{llllrr}
\toprule
\multicolumn{6}{l}{\textbf{Panel C}} \\
Model & \shortstack[l]{T\\ (temperature)} & \shortstack[l]{Layer\\ (index)} & Construction & \shortstack[l]{Angle,\\ general\\ professor vs\\ law\\ professor\\ (degrees)} & \shortstack[l]{Lowest\\ margin 95\%\\ lower\\ bound over\\ roles\\ (cosine)} \\
\midrule
Qwen2.5-7B-Instruct & 1.0 & 19 to 21 & in-domain only (15 questions) & 55.9 to 59.5 & 0.326 \\
Qwen2.5-7B-Instruct & 1.0 & 19 to 21 & out-of-domain only, 75 questions & 54.5 to 58.5 & 0.171 \\
Qwen2.5-7B-Instruct & 1.0 & 19 to 21 & \shortstack[l]{balanced, official (15 in-domain\\ + 75 out-of-domain, 0.5/0.5)} & 54.4 to 58.1 & 0.274 \\
Qwen2.5-7B-Instruct & 1.0 & 19 to 21 & \shortstack[l]{out-of-domain only, 15 questions\\ (matched size)} & 60.4 to 63.7 & 0.160 \\
\addlinespace
Llama-3.1-8B-Instruct & 1.0 & 12 & in-domain only (15 questions) & 45.0 & 0.209 \\
Llama-3.1-8B-Instruct & 1.0 & 12 & out-of-domain only, 75 questions & 42.7 & 0.194 \\
Llama-3.1-8B-Instruct & 1.0 & 12 & \shortstack[l]{balanced, official (15 in-domain\\ + 75 out-of-domain, 0.5/0.5)} & 41.9 & 0.189 \\
Llama-3.1-8B-Instruct & 1.0 & 12 & \shortstack[l]{out-of-domain only, 15 questions\\ (matched size)} & 46.3 & 0.189 \\
\bottomrule
\end{tabular}}
\par\smallskip
{\footnotesize\setlength{\tabcolsep}{3pt}
\begin{tabular}{llllr}
\toprule
\multicolumn{5}{l}{\textbf{Panel D}} \\
Model & \shortstack[l]{T\\ (temperature)} & \shortstack[l]{Layer\\ (index)} & Construction & \shortstack[l]{Cosine to balanced\\ centroid, mean (min\\ to max) over roles\\ (cosine)} \\
\midrule
Qwen2.5-7B-Instruct & 1.0 & 19 to 21 & in-domain only (15 questions) & 0.962 (0.939 to 0.988) \\
Qwen2.5-7B-Instruct & 1.0 & 19 to 21 & out-of-domain only, 75 questions & 0.956 (0.919 to 0.986) \\
Qwen2.5-7B-Instruct & 1.0 & 19 to 21 & \shortstack[l]{balanced, official (15 in-domain\\ + 75 out-of-domain, 0.5/0.5)} & 1 (reference) \\
Qwen2.5-7B-Instruct & 1.0 & 19 to 21 & \shortstack[l]{out-of-domain only, 15 questions\\ (matched size)} & 0.929 (0.868 to 0.972) \\
\addlinespace
Llama-3.1-8B-Instruct & 1.0 & 12 & in-domain only (15 questions) & 0.961 (0.919 to 0.983) \\
Llama-3.1-8B-Instruct & 1.0 & 12 & out-of-domain only, 75 questions & 0.971 (0.938 to 0.985) \\
Llama-3.1-8B-Instruct & 1.0 & 12 & \shortstack[l]{balanced, official (15 in-domain\\ + 75 out-of-domain, 0.5/0.5)} & 1 (reference) \\
Llama-3.1-8B-Instruct & 1.0 & 12 & \shortstack[l]{out-of-domain only, 15 questions\\ (matched size)} & 0.954 (0.917 to 0.973) \\
\bottomrule
\end{tabular}}

\caption{Table~\ref{tab:C1} (continued): Panels C and D, the general
professor vs law professor angle, the lowest margin bound and the cosine to
the injected centroid.}
\end{table*}

\begin{table*}[tp]
\centering
{\footnotesize\setlength{\tabcolsep}{3pt}
\begin{tabular}{lllrrrrr}
\toprule
Cell & Condition & Domain & \shortstack[l]{Answers\\ (count)} & \shortstack[l]{Adherence\\ 0 (\% of\\ answers)} & \shortstack[l]{Adherence\\ 1 (\% of\\ answers)} & \shortstack[l]{Adherence\\ 2 (\% of\\ answers)} & \shortstack[l]{Adherence\\ 3 (\% of\\ answers)} \\
\midrule
Qwen T=1.0 & unsteered & in-domain & 90 & 0.0 & 0.0 & 81.1 & 18.9 \\
Qwen T=1.0 & unsteered & out-of-domain & 450 & 0.0 & 0.4 & 92.2 & 7.3 \\
Qwen T=1.0 & prompted & in-domain & 90 & 0.0 & 0.0 & 20.0 & 80.0 \\
Qwen T=1.0 & prompted & out-of-domain & 450 & 0.0 & 0.2 & 72.9 & 26.9 \\
Qwen T=1.0 & steered & in-domain & 90 & 0.0 & 0.0 & 43.3 & 56.7 \\
Qwen T=1.0 & steered & out-of-domain & 450 & 0.0 & 0.2 & 72.2 & 27.6 \\
Qwen T=1.0 & prompt plus vector & in-domain & 90 & 0.0 & 1.1 & 17.8 & 81.1 \\
Qwen T=1.0 & prompt plus vector & out-of-domain & 450 & 0.0 & 0.0 & 61.1 & 38.9 \\
Qwen T=1.0 & random direction & in-domain & 90 & 0.0 & 0.0 & 71.1 & 28.9 \\
Qwen T=1.0 & random direction & out-of-domain & 450 & 0.0 & 0.2 & 91.6 & 8.2 \\
\midrule
Llama T=0.0 & unsteered & in-domain & 90 & 0.0 & 0.0 & 77.8 & 22.2 \\
Llama T=0.0 & unsteered & out-of-domain & 450 & 0.0 & 0.2 & 94.9 & 4.9 \\
Llama T=0.0 & prompted & in-domain & 90 & 0.0 & 0.0 & 16.7 & 83.3 \\
Llama T=0.0 & prompted & out-of-domain & 450 & 0.0 & 0.2 & 74.2 & 25.6 \\
Llama T=0.0 & steered & in-domain & 90 & 0.0 & 0.0 & 62.2 & 37.8 \\
Llama T=0.0 & steered & out-of-domain & 450 & 0.0 & 0.2 & 89.6 & 10.2 \\
Llama T=0.0 & prompt plus vector & in-domain & 90 & 0.0 & 0.0 & 21.1 & 78.9 \\
Llama T=0.0 & prompt plus vector & out-of-domain & 450 & 0.0 & 0.4 & 63.6 & 36.0 \\
Llama T=0.0 & random direction & in-domain & 90 & 0.0 & 0.0 & 80.0 & 20.0 \\
Llama T=0.0 & random direction & out-of-domain & 450 & 0.0 & 0.4 & 95.3 & 4.2 \\
\bottomrule
\end{tabular}}

\caption{Adherence score distribution (share of answers scoring 0, 1, 2 and 3
on the role's own rubric) by condition and question domain, Qwen $T=1.0$ and
Llama $T=0.0$, 90 in-domain and 450 out-of-domain answers per condition.}
\label{tab:X2}
\end{table*}

\begin{table*}[tp]
\centering
{\footnotesize\setlength{\tabcolsep}{3pt}
\begin{tabular}{llllrl}
\toprule
\multicolumn{6}{l}{\textbf{Panel A}} \\
Model & T & Condition & Domain & $n$ & All six columns \\
\midrule
Qwen & 0.0 & unsteered & out-of-domain & 450 & -- \\
Qwen & 0.0 & steered & out-of-domain & 450 & +0.155 [+0.120, +0.190] \\
Qwen & 0.0 & prompted & out-of-domain & 450 & +0.226 [+0.182, +0.270] \\
Qwen & 0.0 & unsteered & in-domain & 90 & -- \\
Qwen & 0.0 & steered & in-domain & 90 & +0.316 [+0.189, +0.442] \\
Qwen & 0.0 & prompted & in-domain & 90 & +0.453 [+0.353, +0.551] \\
\addlinespace
Qwen & 0.5 & unsteered & out-of-domain & 450 & -- \\
Qwen & 0.5 & steered & out-of-domain & 450 & +0.171 [+0.137, +0.204] \\
Qwen & 0.5 & prompted & out-of-domain & 450 & +0.205 [+0.169, +0.241] \\
Qwen & 0.5 & unsteered & in-domain & 90 & -- \\
Qwen & 0.5 & steered & in-domain & 90 & +0.311 [+0.180, +0.444] \\
Qwen & 0.5 & prompted & in-domain & 90 & +0.511 [+0.398, +0.620] \\
\addlinespace
Qwen & 1.0 & unsteered & out-of-domain & 450 & -- \\
Qwen & 1.0 & steered & out-of-domain & 450 & +0.220 [+0.178, +0.260] \\
Qwen & 1.0 & prompted & out-of-domain & 450 & +0.172 [+0.135, +0.209] \\
Qwen & 1.0 & unsteered & in-domain & 90 & -- \\
Qwen & 1.0 & steered & in-domain & 90 & +0.382 [+0.273, +0.493] \\
Qwen & 1.0 & prompted & in-domain & 90 & +0.544 [+0.438, +0.644] \\
\addlinespace
Qwen & 1.5 & unsteered & out-of-domain & 450 & -- \\
Qwen & 1.5 & steered & out-of-domain & 450 & +0.027 [-0.008, +0.061] \\
Qwen & 1.5 & prompted & out-of-domain & 450 & +0.121 [+0.087, +0.156] \\
Qwen & 1.5 & unsteered & in-domain & 90 & -- \\
Qwen & 1.5 & steered & in-domain & 90 & +0.078 [-0.022, +0.176] \\
Qwen & 1.5 & prompted & in-domain & 90 & +0.240 [+0.124, +0.353] \\
\midrule
\multicolumn{6}{l}{\textit{Success rate difference (pp) [95\% CI]; question-cluster paired bootstrap, 10,000 resamples, seed 42}} \\
Qwen & 0.0 & steered minus unsteered & out-of-domain & 450 & \multicolumn{1}{l}{+13.6 [+9.8, +17.1]} \\
Qwen & 0.0 & prompt plus vector minus prompted & out-of-domain & 450 & \multicolumn{1}{l}{+4.9 [+0.7, +9.1]} \\
Qwen & 0.0 & steered minus unsteered & in-domain & 90 & \multicolumn{1}{l}{+30.0 [+17.8, +42.2]} \\
Qwen & 0.0 & prompt plus vector minus prompted & in-domain & 90 & \multicolumn{1}{l}{+5.6 [-3.3, +14.4]} \\
\addlinespace
Qwen & 0.5 & steered minus unsteered & out-of-domain & 450 & \multicolumn{1}{l}{+16.7 [+12.9, +20.4]} \\
Qwen & 0.5 & prompt plus vector minus prompted & out-of-domain & 450 & \multicolumn{1}{l}{+8.2 [+4.0, +12.4]} \\
Qwen & 0.5 & steered minus unsteered & in-domain & 90 & \multicolumn{1}{l}{+31.1 [+17.8, +44.4]} \\
Qwen & 0.5 & prompt plus vector minus prompted & in-domain & 90 & \multicolumn{1}{l}{-2.2 [-12.2, +7.8]} \\
\addlinespace
Qwen & 1.0 & steered minus unsteered & out-of-domain & 450 & \multicolumn{1}{l}{+20.0 [+15.8, +24.4]} \\
Qwen & 1.0 & prompt plus vector minus prompted & out-of-domain & 450 & \multicolumn{1}{l}{+9.3 [+4.7, +13.8]} \\
Qwen & 1.0 & steered minus unsteered & in-domain & 90 & \multicolumn{1}{l}{+36.7 [+25.6, +47.8]} \\
Qwen & 1.0 & prompt plus vector minus prompted & in-domain & 90 & \multicolumn{1}{l}{+0.0 [-8.9, +8.9]} \\
\addlinespace
Qwen & 1.5 & steered minus unsteered & out-of-domain & 450 & \multicolumn{1}{l}{+2.2 [-1.3, +6.0]} \\
Qwen & 1.5 & steered minus unsteered & in-domain & 90 & \multicolumn{1}{l}{+7.8 [-3.3, +18.9]} \\
\bottomrule
\end{tabular}}

\caption{Specificity by question domain on Qwen for the unsteered, steered
and prompted conditions at every temperature, all six rubrics, as a fraction
(0.220 = 22.0~pp) with 95\% question-cluster intervals and the number of
answers; below, success rate differences (pp, 95\% intervals, question-cluster
paired bootstrap, 10,000 resamples): steered minus unsteered and prompt plus
vector minus prompted, by domain.}
\label{tab:S1q}
\end{table*}

\begin{table*}[tp]
\centering
{\footnotesize\setlength{\tabcolsep}{3pt}
\begin{tabular}{llllrl}
\toprule
\multicolumn{6}{l}{\textbf{Panel A}} \\
Model & T & Condition & Domain & $n$ & All six columns \\
\midrule
Llama & 0.0 & unsteered & out-of-domain & 450 & -- \\
Llama & 0.0 & steered & out-of-domain & 450 & +0.077 [+0.051, +0.104] \\
Llama & 0.0 & prompted & out-of-domain & 450 & +0.206 [+0.164, +0.250] \\
Llama & 0.0 & prompt plus vector & out-of-domain & 450 & +0.319 [+0.276, +0.363] \\
Llama & 0.0 & unsteered & in-domain & 90 & -- \\
Llama & 0.0 & steered & in-domain & 90 & +0.156 [+0.060, +0.251] \\
Llama & 0.0 & prompted & in-domain & 90 & +0.551 [+0.451, +0.651] \\
Llama & 0.0 & prompt plus vector & in-domain & 90 & +0.531 [+0.424, +0.636] \\
\addlinespace
Llama & 0.5 & unsteered & out-of-domain & 450 & -- \\
Llama & 0.5 & steered & out-of-domain & 450 & +0.012 [-0.011, +0.034] \\
Llama & 0.5 & prompted & out-of-domain & 450 & +0.196 [+0.160, +0.232] \\
Llama & 0.5 & prompt plus vector & out-of-domain & 450 & +0.275 [+0.232, +0.318] \\
Llama & 0.5 & unsteered & in-domain & 90 & -- \\
Llama & 0.5 & steered & in-domain & 90 & +0.089 [+0.004, +0.173] \\
Llama & 0.5 & prompted & in-domain & 90 & +0.513 [+0.402, +0.622] \\
Llama & 0.5 & prompt plus vector & in-domain & 90 & +0.449 [+0.336, +0.562] \\
\addlinespace
Llama & 1.0 & unsteered & out-of-domain & 450 & -- \\
Llama & 1.0 & steered & out-of-domain & 450 & +0.058 [+0.031, +0.088] \\
Llama & 1.0 & prompted & out-of-domain & 450 & +0.196 [+0.152, +0.240] \\
Llama & 1.0 & unsteered & in-domain & 90 & -- \\
Llama & 1.0 & steered & in-domain & 90 & -0.044 [-0.140, +0.051] \\
Llama & 1.0 & prompted & in-domain & 90 & +0.389 [+0.260, +0.511] \\
\midrule
\multicolumn{6}{l}{\textit{Success rate difference (pp) [95\% CI]; question-cluster paired bootstrap, 10,000 resamples, seed 42}} \\
Llama & 0.0 & steered minus unsteered & out-of-domain & 450 & \multicolumn{1}{l}{+5.3 [+2.9, +7.8]} \\
Llama & 0.0 & prompt plus vector minus prompted & out-of-domain & 450 & \multicolumn{1}{l}{+10.0 [+6.4, +13.6]} \\
Llama & 0.0 & steered minus unsteered & in-domain & 90 & \multicolumn{1}{l}{+15.6 [+5.6, +25.6]} \\
Llama & 0.0 & prompt plus vector minus prompted & in-domain & 90 & \multicolumn{1}{l}{-4.4 [-12.2, +3.3]} \\
\addlinespace
Llama & 0.5 & steered minus unsteered & out-of-domain & 450 & \multicolumn{1}{l}{+0.0 [-2.4, +2.4]} \\
Llama & 0.5 & prompt plus vector minus prompted & out-of-domain & 450 & \multicolumn{1}{l}{+7.8 [+3.6, +12.0]} \\
Llama & 0.5 & steered minus unsteered & in-domain & 90 & \multicolumn{1}{l}{+6.7 [-2.2, +15.6]} \\
Llama & 0.5 & prompt plus vector minus prompted & in-domain & 90 & \multicolumn{1}{l}{-4.4 [-13.3, +4.4]} \\
\addlinespace
Llama & 1.0 & steered minus unsteered & out-of-domain & 450 & \multicolumn{1}{l}{+1.6 [-1.6, +4.7]} \\
Llama & 1.0 & steered minus unsteered & in-domain & 90 & \multicolumn{1}{l}{-6.7 [-16.7, +3.3]} \\
\bottomrule
\end{tabular}}

\caption{As Table~\ref{tab:S1q}, for Llama, with the prompt-plus-vector
condition included where it was run.}
\label{tab:S1l}
\end{table*}

\begin{table*}[tp]
\centering
{\footnotesize\setlength{\tabcolsep}{3pt}
\begin{tabular}{llrrrl}
\toprule
Grouping & Level & \shortstack[l]{Pairs\\ (count)} & \shortstack[l]{Human\\ agrees with\\ judge\\ (count of\\ decisive\\ pairs)} & \shortstack[l]{Agreement\\ (\% of\\ decisive\\ pairs)} & \shortstack[l]{Exact 95\% CI\\ agreement (\%)} \\
\midrule
Total & all pairs & 50 & 38 of 49 & 77.6 & [63.4, 88.2] \\
\midrule
Domain & in-domain & 25 & 20 of 25 & 80.0 & not computed (small n) \\
Domain & out-of-domain & 25 & 18 of 24 & 75.0 & not computed (small n) \\
\midrule
Role & admin assistant & 9 & 6 of 8 & 75.0 & not computed (small n) \\
Role & athlete & 9 & 8 of 9 & 88.9 & not computed (small n) \\
Role & M\&A lawyer & 8 & 8 of 8 & 100.0 & not computed (small n) \\
Role & law professor & 8 & 6 of 8 & 75.0 & not computed (small n) \\
Role & general professor & 8 & 5 of 8 & 62.5 & not computed (small n) \\
Role & cashier & 8 & 5 of 8 & 62.5 & not computed (small n) \\
\midrule
Condition of the judge-preferred answer & other role & 1 & 1 of 1 & 100.0 & not computed (small n) \\
Condition of the judge-preferred answer & prompted & 19 & 14 of 19 & 73.7 & not computed (small n) \\
Condition of the judge-preferred answer & shared part & 4 & 4 of 4 & 100.0 & not computed (small n) \\
Condition of the judge-preferred answer & role-specific part & 11 & 8 of 10 & 80.0 & not computed (small n) \\
Condition of the judge-preferred answer & steered & 13 & 10 of 13 & 76.9 & not computed (small n) \\
Condition of the judge-preferred answer & unsteered & 2 & 1 of 2 & 50.0 & not computed (small n) \\
\bottomrule
\end{tabular}}

\caption{Blind human check of the judge on 50 answer pairs (Qwen, $T=1.0$).
For each pair a human rater, blind to the condition, chose the answer that
expressed the role better; the table gives agreement with the judge's
preference among decisive pairs, overall with an exact 95\% interval, and by
question domain, role and the condition of the judge-preferred answer (small
groups without intervals).}
\label{tab:M1}
\end{table*}

\begin{figure}[tp]
\centering
\resultsfig{\figcolwidth}{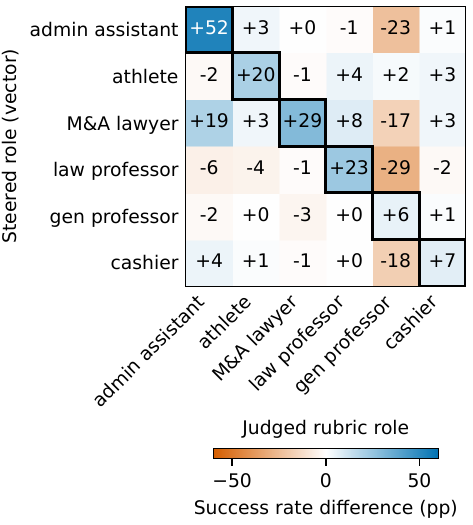}
\caption{Role specificity on Qwen at $T=1.0$ as a $6\times6$ matrix: steered
minus unsteered success rate (pp) for each steered role scored against each of
the six role rubrics. Specificity is the mean of the diagonal (own rubric)
minus the mean of the off-diagonal cells, $+24.7$~pp.}
\label{fig:E2}
\end{figure}

\begin{table*}[tp]
\centering
{\footnotesize\setlength{\tabcolsep}{3pt}
\begin{tabular}{llll}
\toprule
Condition & \shortstack[l]{All questions\\ (540 answers)} & \shortstack[l]{In-domain questions\\ (90 answers)} & \shortstack[l]{Out-of-domain questions\\ (450 answers)} \\
\midrule
\multicolumn{4}{l}{\textit{Qwen2.5-7B-Instruct, T = 1.0: identification rate (\%) [95\% CI]}} \\
unsteered & 17.0 [16.1, 18.1] & 71.1 [61.1, 80.0] & 6.2 [4.2, 8.4] \\
prompted & 37.8 [34.4, 41.1] & 94.4 [88.9, 98.9] & 26.4 [22.7, 30.4] \\
steered & 36.9 [33.5, 40.2] & 94.4 [88.9, 98.9] & 25.3 [21.6, 29.3] \\
prompt plus vector & 49.8 [46.5, 53.3] & 98.9 [96.7, 100.0] & 40.0 [35.8, 44.2] \\
random direction & 16.1 [14.6, 17.4] & 65.6 [55.6, 75.6] & 6.2 [4.2, 8.4] \\
\addlinespace
\multicolumn{4}{l}{\textit{Qwen2.5-7B-Instruct, T = 1.0: difference in identification rate (percentage points) [95\% CI]}} \\
steered minus prompted & $-$0.9 [$-$5.0, +3.1] & 0.0 [$-$6.7, +5.6] & $-$1.1 [$-$5.8, +3.6] \\
prompt plus vector minus prompted & +12.0 [+8.7, +15.4] & +4.4 [+1.1, +8.9] & +13.6 [+9.6, +17.3] \\
\midrule
\multicolumn{4}{l}{\textit{Llama-3.1-8B-Instruct, T = 0.0: identification rate (\%) [95\% CI]}} \\
unsteered & 17.2 [15.9, 18.5] & 68.9 [58.9, 77.8] & 6.9 [4.9, 8.9] \\
prompted & 55.9 [51.9, 59.8] & 95.6 [91.1, 98.9] & 48.0 [43.1, 52.7] \\
steered & 26.5 [24.3, 28.9] & 76.7 [67.8, 85.6] & 16.4 [13.3, 19.6] \\
prompt plus vector & 58.7 [54.6, 62.8] & 98.9 [96.7, 100.0] & 50.7 [45.8, 55.6] \\
random direction & 17.0 [16.1, 18.1] & 71.1 [61.1, 80.0] & 6.2 [4.2, 8.2] \\
\addlinespace
\multicolumn{4}{l}{\textit{Llama-3.1-8B-Instruct, T = 0.0: difference in identification rate (percentage points) [95\% CI]}} \\
steered minus prompted & $-$29.4 [$-$33.7, $-$25.2] & $-$18.9 [$-$28.9, $-$8.9] & $-$31.6 [$-$36.2, $-$26.9] \\
prompt plus vector minus prompted & +2.8 [+0.6, +5.2] & +3.3 [0.0, +7.8] & +2.7 [0.0, +5.3] \\
\bottomrule
\end{tabular}}

\caption{Role identification. The judge is not told the
condition or which role the answer was written for; it sees the question, the
answer and the six roles with their names, descriptions and rubrics in
shuffled order, and picks which of the six roles the answer most sounds like
it came from. Identification rate (percent, 95\% question-cluster bootstrap
intervals, 10,000 resamples; chance 16.7\%) for Qwen at $T=1.0$ (steered 0.3,
prompt plus vector 0.15) and Llama at $T=0.0$ (steered 0.5, prompt plus vector
0.25); 540 answers per condition, 90 to in-domain and 450 to out-of-domain
questions. Below each block, paired differences (pp) over matched questions
with the same resampling scheme. The Qwen steered-minus-prompted row is the
same quantity computed on all answers by the self-naming check. An interval
that includes or touches zero is not read as a difference.}
\label{tab:BID}
\end{table*}

\begin{table*}[tp]
\centering
{\footnotesize\setlength{\tabcolsep}{3pt}
\begin{tabular}{llllrrr}
\toprule
Model & T & \shortstack[l]{Condition (strength),\\ minus prompted} & \shortstack[l]{Answer\\ pairs} & \shortstack[l]{Pooled} & \shortstack[l]{In-domain\\ questions} & \shortstack[l]{Out-of-domain\\ questions} \\
\midrule
Qwen & 0.0 & steered (0.3) & 540 & $-$10.9 [$-$15.2, $-$6.7] & $-$22.2 [$-$33.3, $-$11.1] & $-$8.7 [$-$12.9, $-$4.7] \\
Qwen & 0.0 & prompt plus vector (0.15) & 540 & +5.0 [+1.1, +8.9] & +5.6 [$-$3.3, +14.4] & +4.9 [+0.7, +9.1] \\
Qwen & 0.5 & steered (0.3) & 540 & $-$8.5 [$-$12.6, $-$4.4] & $-$26.7 [$-$37.8, $-$15.6] & $-$4.9 [$-$9.1, $-$0.7] \\
Qwen & 0.5 & prompt plus vector (0.15) & 540 & +6.5 [+2.6, +10.4] & $-$2.2 [$-$12.2, +7.8] & +8.2 [+4.0, +12.4] \\
Qwen & 1.0 & steered (0.3) & 540 & $-$2.6 [$-$7.0, +1.7] & $-$24.4 [$-$35.6, $-$13.3] & +1.8 [$-$2.9, +6.4] \\
Qwen & 1.0 & prompt plus vector (0.15) & 540 & +7.8 [+4.1, +11.5] & +0.0 [$-$8.9, +8.9] & +9.3 [+4.7, +13.8] \\
Qwen & 1.5 & steered (0.1) & 540 & $-$9.3 [$-$13.1, $-$5.4] & $-$20.0 [$-$33.3, $-$6.7] & $-$7.1 [$-$10.9, $-$3.3] \\
Qwen & 1.5 & prompt plus vector & n/a & None feasible & None feasible & None feasible \\
Llama & 0.0 & steered (0.5) & 540 & $-$20.0 [$-$23.3, $-$16.7] & $-$45.6 [$-$55.6, $-$35.6] & $-$14.9 [$-$18.9, $-$11.1] \\
Llama & 0.0 & prompt plus vector (0.25) & 540 & +7.6 [+4.4, +10.9] & $-$4.4 [$-$12.2, +3.3] & +10.0 [+6.4, +13.6] \\
Llama & 0.5 & steered (0.2) & 540 & $-$19.4 [$-$23.1, $-$15.7] & $-$46.7 [$-$57.8, $-$35.6] & $-$14.0 [$-$17.6, $-$10.4] \\
Llama & 0.5 & prompt plus vector (0.15) & 540 & +5.7 [+2.2, +9.3] & $-$4.4 [$-$13.3, +4.4] & +7.8 [+3.6, +12.0] \\
Llama & 1.0 & steered (0.2) & 540 & $-$19.4 [$-$23.5, $-$15.2] & $-$47.8 [$-$58.9, $-$35.6] & $-$13.8 [$-$18.7, $-$9.1] \\
Llama & 1.0 & prompt plus vector & n/a & None feasible & None feasible & None feasible \\
\bottomrule
\end{tabular}}

\caption{Steered minus prompted, and prompt plus vector minus prompted,
success rate (pp, 95\% question-cluster intervals in brackets) over all 540
answer pairs, the 90 in-domain and the 450 out-of-domain, by model and
temperature; each row label gives the condition and its strength in
reference-norm units. ``None feasible'': calibration ran and no strength kept
mean coherence within the floor.}
\label{tab:F2}
\end{table*}

\begin{table*}[tp]
\centering
{\footnotesize\setlength{\tabcolsep}{3pt}
\begin{tabular}{llll}
\toprule
Condition & \shortstack[l]{All questions\\ (540 answers)} & \shortstack[l]{In-domain questions\\ (90 answers)} & \shortstack[l]{Out-of-domain questions\\ (450 answers)} \\
\midrule
\multicolumn{4}{l}{\textit{Qwen2.5-7B-Instruct, T = 1.0: answers that name their own role (count and share)}} \\
unsteered & 0 (0.0\%) & 0 (0.0\%) & 0 (0.0\%) \\
prompted & 1 (0.2\%) & 0 (0.0\%) & 1 (0.2\%) \\
steered & 1 (0.2\%) & 1 (1.1\%) & 0 (0.0\%) \\
prompt plus vector & 3 (0.6\%) & 0 (0.0\%) & 3 (0.7\%) \\
random direction & 0 (0.0\%) & 0 (0.0\%) & 0 (0.0\%) \\
\addlinespace
\multicolumn{4}{l}{\textit{Llama-3.1-8B-Instruct, T = 0.0: answers that name their own role (count and share)}} \\
unsteered & 0 (0.0\%) & 0 (0.0\%) & 0 (0.0\%) \\
prompted & 130 (24.1\%) & 11 (12.2\%) & 119 (26.4\%) \\
steered & 0 (0.0\%) & 0 (0.0\%) & 0 (0.0\%) \\
prompt plus vector & 132 (24.4\%) & 12 (13.3\%) & 120 (26.7\%) \\
random direction & 0 (0.0\%) & 0 (0.0\%) & 0 (0.0\%) \\
\midrule
\multicolumn{4}{l}{\textit{Llama-3.1-8B-Instruct, T = 0.0: difference in identification rate (percentage points) [95\% CI]}} \\
\multicolumn{4}{l}{Steered minus prompted} \\
\quad all answers & \shortstack[l]{$-$29.44 [$-$33.70, $-$25.19]\\ (540 pairs)} & \shortstack[l]{$-$18.89 [$-$28.89, $-$8.89]\\ (90 pairs)} & \shortstack[l]{$-$31.56 [$-$36.22, $-$26.89]\\ (450 pairs)} \\
\quad self-naming answers excluded & \shortstack[l]{$-$11.71 [$-$15.99, $-$7.52]\\ (410 pairs)} & \shortstack[l]{$-$17.72 [$-$29.11, $-$6.33]\\ (79 pairs)} & \shortstack[l]{$-$10.27 [$-$15.17, $-$5.44]\\ (331 pairs)} \\
\multicolumn{4}{l}{Prompt plus vector minus prompted} \\
\quad all answers & \shortstack[l]{+2.78 [+0.56, +5.19]\\ (540 pairs)} & \shortstack[l]{+3.33 [0.00, +7.78]\\ (90 pairs)} & \shortstack[l]{+2.67 [0.00, +5.33]\\ (450 pairs)} \\
\quad self-naming answers excluded & \shortstack[l]{+3.70 [+0.74, +6.70]\\ (405 pairs)} & \shortstack[l]{+3.85 [0.00, +8.97]\\ (78 pairs)} & \shortstack[l]{+3.67 [0.00, +7.28]\\ (327 pairs)} \\
\bottomrule
\end{tabular}}

\caption{Answers that name their own role outright (``self-naming'') and
role identification without them. Top: number (and share) of answers
matching a text pattern for self-naming, by condition, for Qwen at $T=1.0$
and Llama at $T=0.0$, over all 540 answers, the 90 in-domain and the 450
out-of-domain. The pattern is a first-person marker (for example ``As a'',
``I'm a'', ``my role as''), at most four words, then a term for the answer's
own role; a spot check found a few false positives on Qwen (analogies such as
``just as an athlete'') and a few misses on Llama (for example ``As your M\&A
lawyer''; adding that marker gives 133 prompted answers), so the counts are
approximate. Bottom: Llama at $T=0.0$, differences in role identification
rate (pp, 95\% question-cluster intervals) on all answers and with
self-naming answers excluded; a question pair is kept only if neither of its
two answers is self-naming (331 of 450 out-of-domain pairs for steered minus
prompted, 327 for prompt plus vector minus prompted), so the exclusion
comparison is conditional on answers that do not name the role. Intervals
that include or touch zero are not read as differences.}
\label{tab:SN}
\end{table*}

\begin{table*}[tp]
\centering
{\footnotesize\setlength{\tabcolsep}{3pt}
\begin{tabular}{llrll}
\toprule
\multicolumn{5}{l}{\textbf{Panel A}} \\
Model & \shortstack[l]{Temperature\\ (T)} & \shortstack[l]{Vector strength\\ (reference-norm\\ units)} & \shortstack[l]{All pairs success rate    (pp)\\ {}[95\% CI] (pairs)} & \shortstack[l]{Excluding length-limit pairs\\ success rate (pp) [95\% CI]\\ (pairs)} \\
\midrule
Qwen & 0.0 & 0.15 & +4.9 [+0.7, +9.1] (n=450) & +5.7 [+1.6, +9.9] (n=439) \\
Qwen & 0.5 & 0.15 & +8.2 [+4.0, +12.4] (n=450) & +8.8 [+4.5, +13.2] (n=443) \\
Qwen & 1.0 & 0.15 & +9.3 [+4.7, +13.8] (n=450) & +10.9 [+6.4, +15.2] (n=442) \\
Llama & 0.0 & 0.25 & +10.0 [+6.4, +13.6] (n=450) & +9.8 [+6.3, +13.4] (n=448) \\
Llama & 0.5 & 0.15 & +7.8 [+3.6, +12.0] (n=450) & +7.8 [+3.6, +12.0] (n=449) \\
\bottomrule
\end{tabular}}
\par\smallskip
{\footnotesize\setlength{\tabcolsep}{3pt}
\begin{tabular}{llrll}
\toprule
\multicolumn{5}{l}{\textbf{Panel B}} \\
Model & \shortstack[l]{Temperature\\ (T)} & \shortstack[l]{Vector strength\\ (reference-norm\\ units)} & \shortstack[l]{Vector answer no longer\\ than prompted success rate\\ (pp) [95\% CI] (pairs)} & \shortstack[l]{Blind\\ identification,\\ accuracy\\ (pp) [95\% CI]} \\
\midrule
Qwen & 0.0 & 0.15 & +8.5 [+1.4, +15.5] (n=142) & not run \\
Qwen & 0.5 & 0.15 & +14.5 [+8.1, +21.3] (n=152) & not run \\
Qwen & 1.0 & 0.15 & +12.4 [+4.9, +19.9] (n=145) & +13.6 [+9.6, +17.3] \\
Llama & 0.0 & 0.25 & +1.8 [$-$3.6, +7.2] (n=168) & not run \\
Llama & 0.5 & 0.15 & +6.3 [+1.4, +11.4] (n=221) & not run \\
\bottomrule
\end{tabular}}
\par\smallskip
{\footnotesize\setlength{\tabcolsep}{3pt}
\begin{tabular}{llrrr}
\toprule
\multicolumn{5}{l}{\textbf{Panel C}} \\
Model & \shortstack[l]{Temperature\\ (T)} & \shortstack[l]{Vector strength\\ (reference-norm\\ units)} & \shortstack[l]{Mean\\ coherence,\\ prompted /\\ vector\\ (judge score\\ 0-100)} & \shortstack[l]{Length-limit\\ share,\\ prompted /\\ vector (\% of\\ answers)} \\
\midrule
Qwen & 0.0 & 0.15 & 96.3 / 92.9 & 0.0 / 2.4 \\
Qwen & 0.5 & 0.15 & 96.1 / 93.2 & 0.0 / 1.6 \\
Qwen & 1.0 & 0.15 & 95.2 / 91.2 & 0.0 / 1.8 \\
Llama & 0.0 & 0.25 & 95.3 / 93.7 & 0.2 / 0.2 \\
Llama & 0.5 & 0.15 & 95.5 / 94.4 & 0.0 / 0.2 \\
\bottomrule
\end{tabular}}

\caption{Length checks for prompt plus vector minus prompted out of domain
(pp, 95\% question-cluster intervals, pair counts), every setting where prompt
plus vector was run. Panel A: all pairs, and excluding pairs in which either
answer hit the length limit. Panel B: pairs in which the vector answer is no
longer than the prompted one (a sensitivity check, since the subset is
selected on length), and role identification where run. Panel C:
mean coherence and share of answers at the length limit, prompted / prompt
plus vector.}
\label{tab:G1}
\end{table*}

\begin{table*}[tp]
\centering
\small
\setlength{\tabcolsep}{3pt}   %
\begin{tabular}{lrlr}
\toprule
Conversations included & \shortstack[l]{Conversations\\ (count)} & \shortstack[l]{Difference in success rate, turns 26-30\\ (pp) [95\% CI]} & \shortstack[l]{Wilcoxon\\ p} \\
\midrule
All 18 conversations & 18 & +13.3 [+2.2, +24.4] & 0.044 \\
Without the general professor & 15 & +16.0 [+5.3, +26.7] & 0.017 \\
Without turns cut off at 512 tokens & 14 & +14.6 [+3.6, +26.4] & 0.044 \\
The 12 conversations never used for calibration & 12 & +20.0 [+6.7, +33.3] & 0.028 \\
\bottomrule
\end{tabular}

\caption{Late-turn robustness on Qwen at $T=1.0$: steered minus prompted
success rate over turns 26 to 30 (pp, 95\% conversation bootstrap interval,
tie-robust Wilcoxon signed-rank $p$) on all 18 conversations ($+13.3$),
without the general professor ($+16.0$), without turns cut off at the
512-token limit ($+14.6$), and on the 12 conversations never used for
calibration ($+20.0$).}
\label{tab:H3}
\end{table*}

\begin{figure*}[tp]
\centering
\begin{minipage}[t]{0.48\textwidth}\centering
\resultsfig{\linewidth}{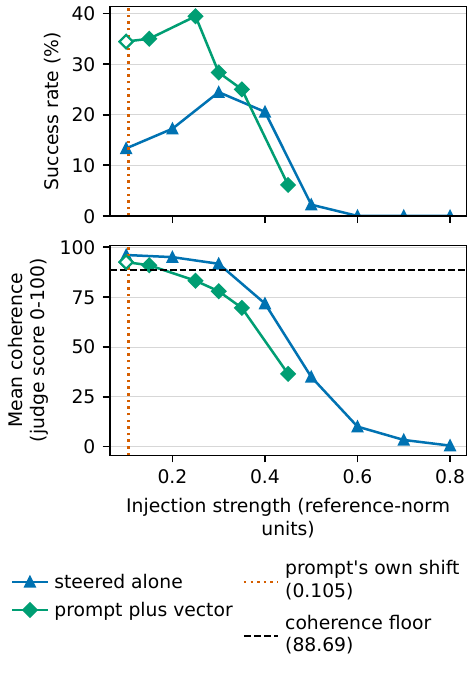}
\caption{Calibration on Qwen at $T=1.0$: success rate and mean coherence on
the calibration questions by injection strength, for steering alone and prompt
plus vector; the chosen strength is the most successful one that keeps mean
coherence above the coherence floor.}
\label{fig:G2}
\end{minipage}\hfill
\begin{minipage}[t]{0.48\textwidth}\centering
\includegraphics[width=\linewidth,height=0.8\textheight,keepaspectratio]{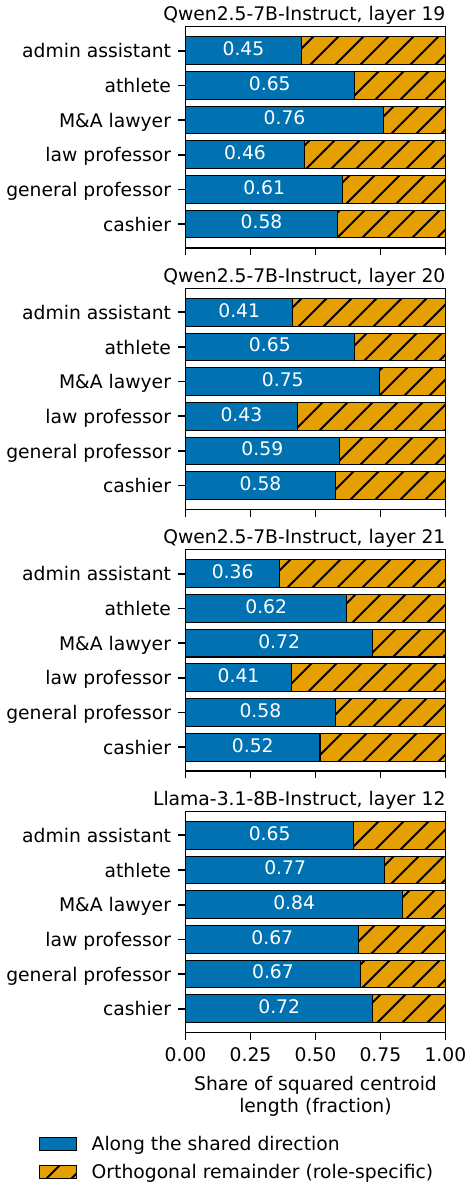}
\caption{As Figure~\ref{fig:J1} (left), with one panel per injection layer:
Qwen layers 19, 20 and 21, and Llama layer 12.}
\label{fig:J1b}
\end{minipage}
\end{figure*}

\begin{table}[tp]
\centering
{\footnotesize\setlength{\tabcolsep}{3pt}
\begin{tabular}{lrrr}
\toprule
Model & \shortstack[l]{Site-by-strength\\ cells (count)} & \shortstack[l]{Cells with\\ mean\\ adherence\\ above\\ unsteered,\\ all answers\\ (count)} & \shortstack[l]{Cells with\\ mean\\ adherence\\ above\\ unsteered,\\ in-domain\\ answers\\ (count)} \\
\midrule
Qwen & 70 & 37 of 70 & 40 of 70 \\
Llama & 103 & 0 of 103 & 0 of 103 \\
\bottomrule
\end{tabular}}

\caption{Exploratory sweep over injection sites and strengths: number of
site-by-strength combinations in which mean adherence exceeds the unsteered
level, over all answers and over in-domain answers; Qwen 37 of 70, Llama 0 of
103.}
\label{tab:I2c}
\end{table}

\begin{table*}[tp]
\centering
{\footnotesize\setlength{\tabcolsep}{3pt}
\begin{tabular}{lrrrr}
\toprule
\multicolumn{5}{l}{\textbf{Panel A}} \\
Model, site & \shortstack[r]{Strength\\ (ref.-norm\\ units)} & \shortstack[r]{French\\ share minus\\ unsteered\\ (original\\ detector)} & \shortstack[r]{Length ratio\\ (tokens)} & \shortstack[r]{Length ratio\\ (characters)} \\
\midrule
Qwen, layers 19-21 & 0.2 & +1.000 & 1.61 & 1.12 \\
 & 0.3 & +0.967 & 1.74 & 1.14 \\
\addlinespace
Llama, layer 12 & 0.2 & +0.600 & 2.06 & 1.56 \\
\addlinespace
Llama, layer 15 & 0.2 & +0.500 & 2.20 & 1.61 \\
\bottomrule
\end{tabular}}
\par\smallskip
{\footnotesize\setlength{\tabcolsep}{3pt}
\begin{tabular}{lrcccc}
\toprule
\multicolumn{6}{l}{\textbf{Panel B}} \\
Model, site & \shortstack[r]{Strength\\ (ref.-norm\\ units)} & \shortstack[r]{Clean\\ French and\\ coherent\\ share\\ (amended)} & \shortstack[c]{Verdict (i)\\ original\\ rule, tokens} & \shortstack[c]{Verdict (ii)\\ original\\ rule,\\ characters} & \shortstack[c]{Verdict (iii)\\ amended\\ rule} \\
\midrule
Qwen, layers 19-21 & 0.2 & 0.767 & fail & pass & pass \\
 & 0.3 & 0.233 &  &  &  \\
\addlinespace
Llama, layer 12 & 0.2 & 0.000 & fail & fail & fail \\
\addlinespace
Llama, layer 15 & 0.2 & 0.000 & fail & fail & fail \\
\bottomrule
\end{tabular}}

\caption{French positive control: a language vector (system prompt ``Always
answer in French.'' against the native default side) injected at the
role-vector sites, 30 calibration questions, $T=1.0$, strengths 0.1 to 0.8
(rows show the strengths at which the original share condition is met).
\emph{Original rule} (committed before any run): the vector steers if, at some
strength, the share of answers detected as French is at least 50 pp above the
unsteered share and the mean answer length is within 50\% of the unsteered
mean answer length. The original rule did not name a length unit; the script
measured tokens, and French text takes more tokens than English text for the
same content, which inflates the token ratio (verdict i); verdict (ii)
measures length in characters. \emph{Amended rule} (verdict iii; amendment
dated 2026-10-04 20:28, committed after the original-definition Llama result
had been seen at 14:21 and the garbled Llama output had been inspected): an
answer is clean French if at least 98\% of its letters are Latin script, its
share of French function words is above 0.12 and above its share of English
function words, and the median of three coherence judge trials is at least 75;
the vector steers if the clean-French-and-coherent share is at least 50 pp
above the unsteered share at some strength. The amended rule is the primary
reading; all three readings are shown. Verdicts are per model and site over
all strengths. Printed as two panels.}
\label{tab:I2b}
\end{table*}

\end{document}